\documentclass[10pt,twocolumn,letterpaper]{article}
\usepackage[accsupp]{axessibility}
\usepackage[algorithms]{wacv}
\usepackage{times}
\usepackage{epsfig}
\usepackage{graphicx}
\usepackage{amsmath}
\usepackage{amssymb}
\usepackage{booktabs}

\usepackage{multirow}
\usepackage[table]{xcolor}
\usepackage{arydshln}
\usepackage{colortbl}
\usepackage{anyfontsize}
\usepackage{overpic}
\usepackage{makecell}

\DeclareUnicodeCharacter{FFFC}{ }

\usepackage[pagebackref,breaklinks,colorlinks]{hyperref}

\def\wacvPaperID{1380} 
\def\confName{WACV}
\def\confYear{2024}

\begin{document}

\title{4K-Resolution Photo Exposure Correction at 125 FPS with $\sim$8K Parameters}

\newcommand*\samethanks[1][\value{footnote}]{\footnotemark[#1]}
\author{
		Yijie Zhou$^{1}$
        ~~~~
		Chao Li$^{1}$
        ~~~~        
        Jin Liang$^{1}$
		~~~~
  Tianyi Xu$^{1}$
		~~~~
            Xin Liu$^{2,3,}$\
        ~~~~
            Jun Xu$^{1,4,}$\thanks{Corresponding author: csjunxu@nankai.edu.cn.}\\
        $^{1}$Nankai University~~
        $^{2}$Tianjin University~~
        $^{3}$Lappeenranta-Lahti University of Technology\\
        $^{4}$Guangdong Provincial Key Laboratory of Big Data Computing, CUHK (Shenzhen)
	}

\maketitle
\thispagestyle{empty}

\begin{abstract}
The illumination of improperly exposed photographs has been widely corrected using deep convolutional neural networks or Transformers. Despite with promising performance, these methods usually suffer from large parameter amounts and heavy computational FLOPs on high-resolution photographs. In this paper, we propose extremely light-weight (with only $\sim$8K parameters) Multi-Scale Linear Transformation (MSLT) networks under the multi-layer perception architecture, which can process 4K-resolution sRGB images at 125 Frame-Per-Second (FPS) by a Titan RTX GPU. Specifically, the proposed MSLT networks first decompose an input image into high and low frequency layers by Laplacian pyramid techniques, and then sequentially correct different layers by pixel-adaptive linear transformation, which is implemented by efficient bilateral grid learning or $1\times1$ convolutions. Experiments on two benchmark datasets demonstrate the efficiency of our MSLTs against the state-of-the-arts on photo exposure correction. Extensive ablation studies validate the effectiveness of our contributions. The code is available at \url{https://github.com/Zhou-Yijie/MSLTNet}.
\end{abstract}

\section{Introduction}
\label{sec:intro}
The prevalence of smartphones with cameras encourages people to take snapshots of their daily life like photographers. However, inaccurate setting of shutter speed, focal-aperture ratio and/or ISO value may bring improper exposure to the captured photographs with degradation on visual quality~\cite{afifi2021learning}. To adjust the photo exposure properly for visually appealing purpose, it is essential to develop efficient exposure correction methods for edge devices.

In last decades, low-light enhancement methods~\cite{chen2018learning,ren2019low,jiang2021enlightengan} and over exposure correction methods~\cite{cao2020over, abebe2021content} have been proposed to adjust the brightness of under-exposed and over-exposed images, respectively. However, low-light enhancement methods could hardly correct over-exposed images while over-exposure correction methods would fail on under-exposed images~\cite{afifi2021learning}. High dynamic range (HDR) tone-mapping methods~\cite{eilertsen2017hdr,Liang_2018_CVPR,liu2020single,2021How} can also adjust improper illumination of the contents to some extent, but mainly enhance local details in improperly-exposed areas along with dynamic range reduction. In the end, all these methods are not suitable for exposure correction, which requires globally adjustment on improper exposure in images.

\begin{figure}[t]
\centering	\includegraphics[width=0.23\textwidth]{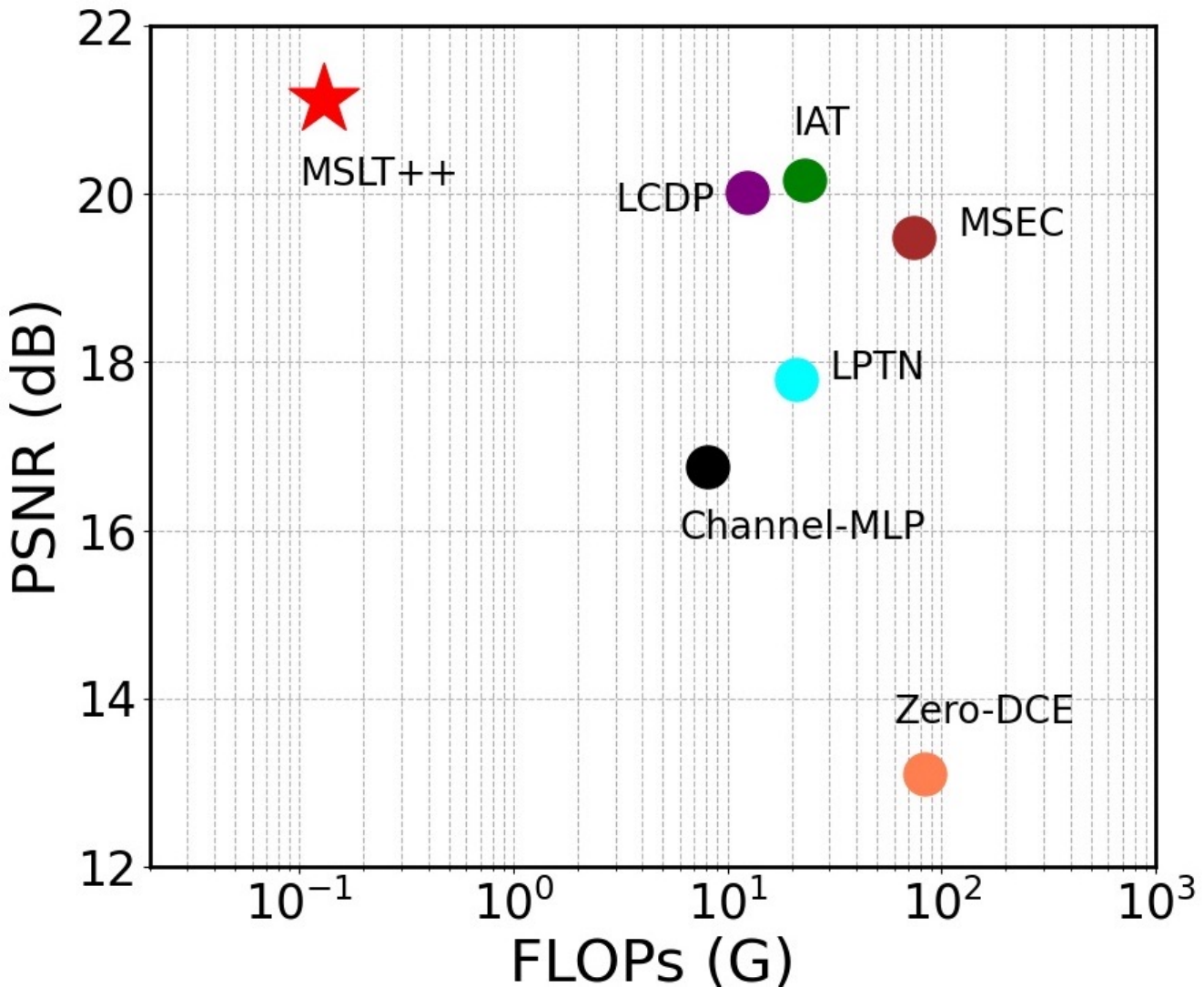}
\includegraphics[width=0.23\textwidth]{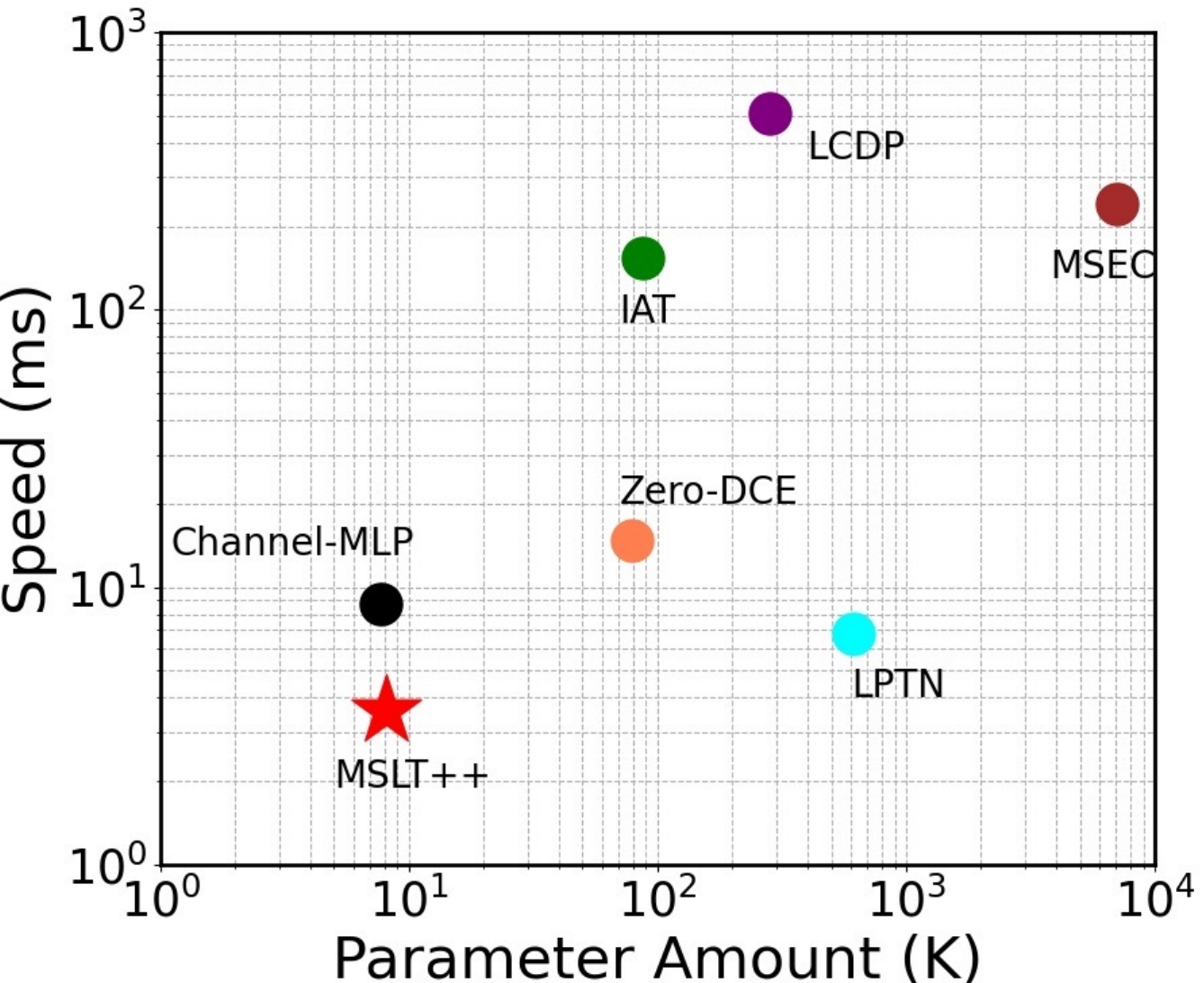}
\caption{\textbf{Comparison of the proposed MSLT++ and state-of-the-art exposure correction methods} on the ME dataset~\cite{afifi2021learning}. Left: comparison of PSNR results and computational costs (FLOPs). Right: comparison of speed (inference time on a $1024\times1024$ sRGB image) and parameter amounts.}
\vspace{-6mm}
\label{fig:compare}
\end{figure}

Recently, there emerges several exposure correction methods based on Convolutional Neural Networks (CNN)~\cite{afifi2021learning} or Transformer~\cite{90K}. For example, Multi-Scale Exposure Correction (MSEC)~\cite{afifi2021learning} performs hierarchical exposure correction with Laplacian pyramid techniques~\cite{burt1987laplacian,denton2015deep,lai2017deep} and the UNet architecture~\cite{2015unet}.
Later, the work of~\cite{wang2022local} exploits the Local Color Distributions Prior (LCDP) to locate and enhance the improperly exposed region.
The attention-based Illumination Adaptive
Transformer (IAT)~\cite{90K} estimates the parameters related to the Image Signal Processor (ISP) under the Transformer architecture~\cite{atten2017}.
%
Despite with promising performance, these exposure correction CNNs or Transformers are limited by huge parameter amounts and computational costs~\cite{afifi2021learning,90K}.

To produce visually pleasing results while still improving the model efficiency, in this paper, we propose extremely light-weight Multi-Scale Linear Transformation (MSLT) networks for high-resolution image exposure correction.
Specifically, we first decompose the input image into high-frequency and low-frequency layers via Laplacian pyramid techniques~\cite{burt1987laplacian,denton2015deep,lai2017deep} to perform coarse-to-fine exposure correction.
We then design simple linear transformation networks to progressively correct these layers, consuming small parameter amounts and computational costs.
For the low-frequency layer, we adopt the bilateral grid learning (BGL) framework~\cite{gharbi2017deep,zheng2021ultra,xu2021bilateral} to learn pixel-wise affine transformation between improper and proper exposed image pairs.
To learn context-aware transformation coefficients in BGL, we propose a parameter-free Context-aware Feature Decomposition (CFD) module and extend it for multi-scale affine transformation.
For the high-frequency layers, we simply learn pixel-wise correction masks by two channel-wise $1\times1$ convolutional layers.

Benefited by using channel-wise multi-layer perception (MLP) for coarse-to-fine exposure correction, our largest network MSLT++ has 8,098 parameters, while requiring only 0.14G and 3.67ms to process a $1024\times1024\times3$ image with a RTX GPU. As a comparison, the parameter amounts of CNN-based MSEC~\cite{afifi2021learning}, LCDP~\cite{wang2022local} and transformer-based IAT~\cite{90K} are $\sim$7,015K, $\sim$282K and $\sim$86.9K, respectively, while the corresponding FLOPs/speed are 73.35G/240.46ms, 17.33G/507.67ms and 22.96G/153.96ms, respectively.
Experiments on two benchmark datasets~\cite{afifi2021learning,cai2018learning} show that our MSLTs achieve better quantitative and qualitative performance than state-of-the-art exposure correction methods. A quick glimpse of comparison on the ME dataset is shown in Figure~\ref{fig:compare}.

Our main contributions are summarized as follows:
\begin{itemize}
\item We develop Multi-Scale Linear Transformation networks with at most 8,098 parameters, which run at most 125 FPS on 4K-resolution ($3840\times2160\times3$) images with effective exposure correction performance.

\item To accelerate the multi-scale decomposition, we design a bilateral grid network (BGN) to pixel-wisely correct the exposure of low-frequency layer. Here, we implement BGN via a channel-wise MLP, rather than CNNs or Transformers, to endow our MSLTs with small parameter amounts and computational costs.
    
\item We propose a Context-aware Feature Decomposition (CFD) module to learn hierarchical transformation coefficients in our BGN for effective exposure correction.
    
\end{itemize}

\section{Related Work}
\label{sec:related}
\subsection{Image Exposure Correction Methods}
The exposure correction task is similar but different to the tasks of low-light image enhancement~\cite{chen2018learning,jiang2021enlightengan}, over-exposure correction~\cite{cao2020over,abebe2021content}, and HDR tone mapping~\cite{2021How,liu2020single,eilertsen2017hdr,Liang_2018_CVPR}.
As far as we know, the work of MSEC~\cite{afifi2021learning} is among the first deep learning based method for exposure correction.
It decomposes an image into high-frequency and low-frequency parts, and progressively corrects the exposure errors.
However, MSEC has over 7M
parameters and is not efficient enough on high-resolution images.
%
The Local Color Distributions Prior (LCDP)~\cite{wang2022local} exploits the local color distributions to uniformly tackle the under-exposure and over-exposure, with about 282K parameters and requires huge computational costs, \eg, 17.33G FLOPs, to process a $1024\times1024\times3$ image.
The Transformer based Illumination-Adaptive-Transformer (IAT)~\cite{90K} has about 86.9K parameters, but suffering from large computational costs and slow inference speed on high-resolution images.

In this paper, we propose light-weight and efficient Multi-Scale Linear Transformation (MSLT) networks, which at most have 8,098 parameters and run at 125 FPS to correct 4K resolution images with improper exposures.

\subsection{Image Processing MLPs}
Multi-layer perceptions (MLPs)~\cite{1988Learning} play an important role in visual tasks before the prosperity of convolutional neural networks (CNNs) and Transformers. 
MLP based networks have attracted the attention of researchers again for its simplicity.
The method of MLP-Mixer~\cite{MLPMixer2021} is a purely MLP-based network without convolutions or self-attention.
Later, ResMLP~\cite{touvron2022resmlp} is proposed using only linear layers and GELU non-linearity.
The work of gMLP~\cite{gmlp2021} utilizes MLPs with gating to achieve comparable results with Transformers~\cite{DeiT,VIT} on image classification~\cite{imagenet}.
Ding et al.~\cite{ding2021repmlp} proposed a re-parameterization technique to boost the capability of MLP on image classification.
The recently developed MAXIM~\cite{tu2022maxim} is a multi-axis MLP based network for general image processing tasks.
In this paper, we develop an extremely efficient exposure correction network, which mainly utilizes channel-wise (not spatial-wise) MLPs to globally perceive the exposure information of the image.

\subsection{Light-weight Image Enhancement Networks}

In pursuit for light-weight and efficient models, one naive way is to apply the model at a low-resolution input and then resize the output into high-resolutions.
But the high-frequency details would be lost.
To this end, the Laplacian Pyramid decomposition ~\cite{burt1987laplacian,afifi2021learning} is used to preserve high-frequency information.
A further approach is to learn an approximate operator at downsampled inputs and then apply this operator to the original image~\cite{chen2016bilateral,gharbi2017deep,ma2019deep}.
Such approximate operators are usually simple and efficient.
Later, this approximation insight is also studied by bilateral grid learning~\cite{chen2007real}, to accelerate diverse image processing methods on the tasks of image enhancement~\cite{gharbi2017deep}, image dehazing~\cite{zheng2021ultra}, and stereo matching~\cite{xu2021bilateral}, \etc.

\begin{figure*}[h]
    \begin{center}
\includegraphics[width=1\textwidth]{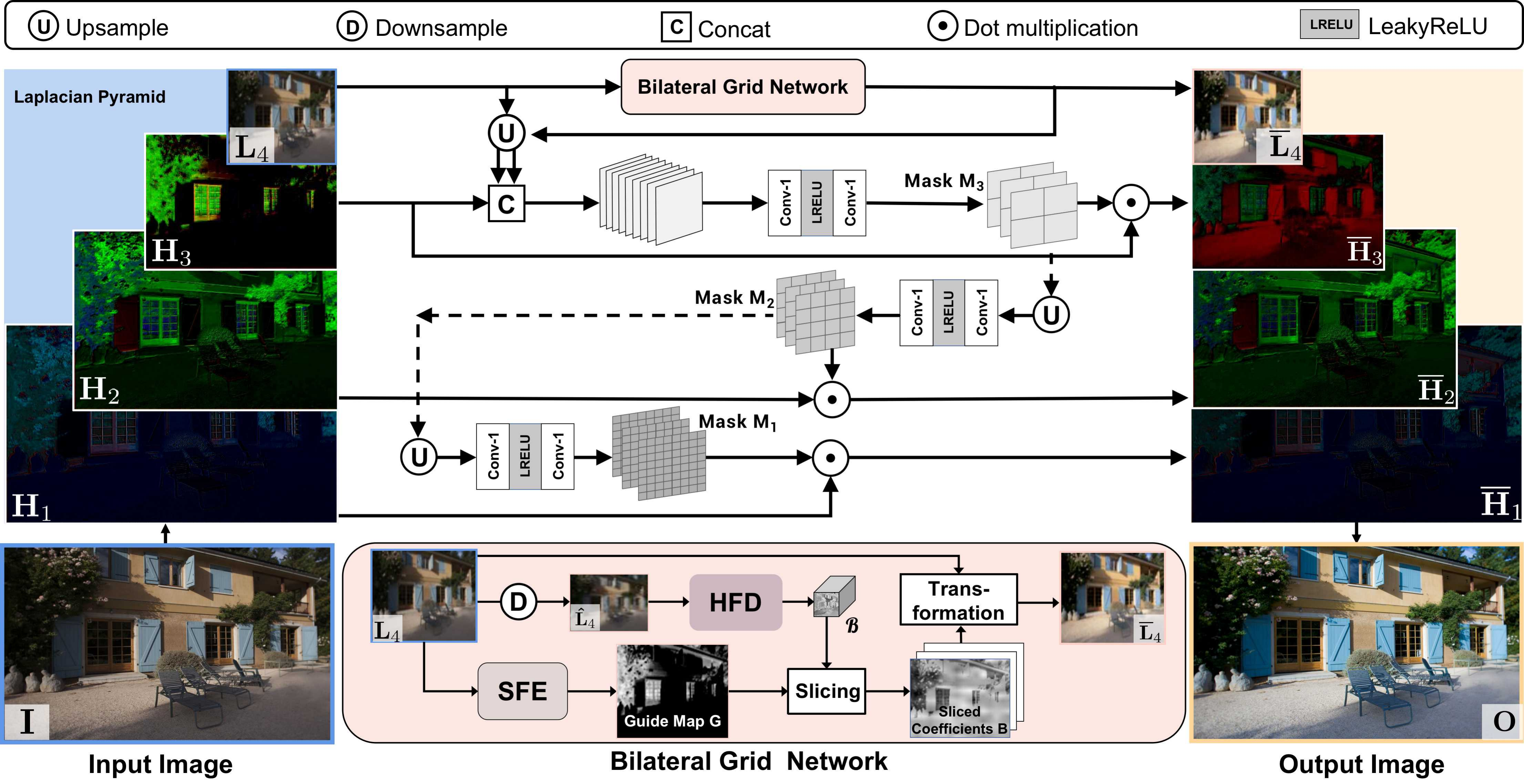}
\vspace{-2mm} 
    \caption{\textbf{Overview of our Multi-Scale Linear Transformation (MSLT) network} with $n=4$.
    Given an input image $\mathbf{I}\in\mathbb{R}^{H\times W\times3}$ with improper exposure, our MSLT firstly decomposes the image $\mathbf{I}$ into high frequency layers $\{\mathbf{H}_i\in\mathbb{R}^{\frac{H}{2^{i-1}}\times\frac{W}{2^{i-1}}\times3}|i=1,2,3\} $ and a low frequency layer $\mathbf{L}_{4}$ by Laplacian pyramid decomposition.
    The $\mathbf{L}_{4}$ is corrected by the proposed Bilateral Grid Network: 1) the $\mathbf{L}_{4}$ is input to Self-modulated Feature Extraction (SFE) module to obtain a guidance map $\mathbf{G}$, 2) the $\mathbf{L}_{4}$ is downsampled to $\mathbf{\hat{L}}_{4}$ of size $48\times48\times3$, which is used to learn the 3D bilateral grid of affine coefficients $\mathcal{B}$ by the Hierarchical Feature Decomposition (HFD) module, 3) with the guidance map $\mathbf{G}$, the coefficients $\mathcal{B}$ are sliced to produce a 2D grid of coefficients $\mathbf{B}$, which is used to pixel-wisely correct the $\mathbf{L}_{4}$.
    The high frequency layers $\{\mathbf{H}_{i}|i=1,2,3\} $ are corrected by learning corresponding masks via two 1$\times$1 convolutions.
    Finally, the corrected low/high-frequency layers are reconstructed to output the exposure corrected image $\mathbf{O}$.
    The SFE and HFD modules are detailed in Figure~\ref{fig:HFD}. }
    \label{fig:MSLT}
    \end{center}
    \vspace{-7mm}
    
\end{figure*}

In this paper, we design light-weight and efficient image exposure correction networks with Laplacian pyramid technique and bilateral grid learning framework. Differently, our bilateral grid network is purely implemented by channel-wise MLP, consuming much less parameters and computational costs than CNNs and Transformers.

\section{Proposed Method}\label{sec:method}

\subsection{Network Overview}
\label{sec:overview}


As illustrated in Figure \ref{fig:MSLT}, our Multi-Scale Linear Transformation (MSLT) network for exposure correction is consisted of four close-knit parts introduced as follows.

\noindent
\textbf{Multi-Scale Image Decomposition}.
As suggested in~\cite{afifi2021learning}, the coarse-to-fine architecture is effective for the exposure correction task.
Given an input image $\mathbf{I}\in\mathbb{R}^{H\times W\times3}$, we employ the Laplacian pyramid technique~\cite{burt1987laplacian} to decompose the image $\mathbf{I}$ into a sequence of $n-1$ high-frequency layers $\{ {\mathbf{H}_i\in\mathbb{R}^{\frac{H}{2^{i-1}}\times\frac{W}{2^{i-1}}\times3}|i=1,...,n-1}\}$ and one low-frequency layer $\mathbf{L}_{n}\in\mathbb{R}^{\frac{H}{2^{n-1}}\times\frac{W}{2^{n-1}}\times3}$.

\begin{figure*}[htp]
    \centering
     \vspace{-2mm}
          \begin{overpic}[width=1\textwidth]{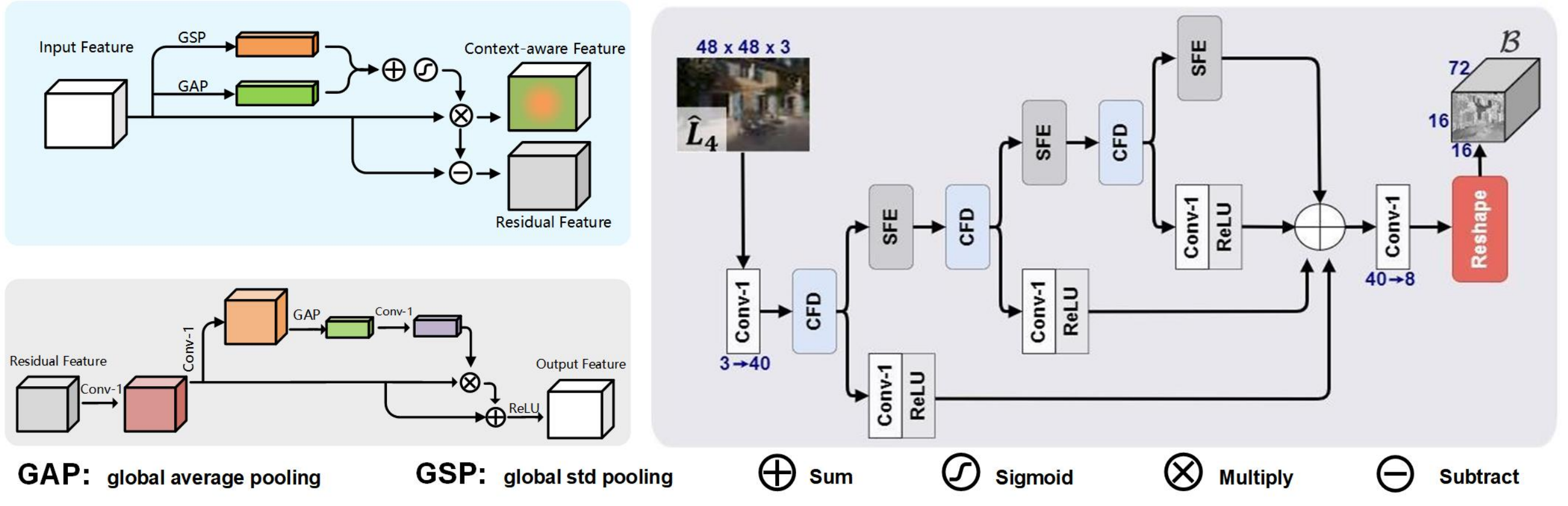}
            \put(0,33){(a) Context-aware Feature Decomposition (CFD)}
            \put(0,14.85){(b) Self-modulated Feature Extraction (SFE)}
            \put(42,33){(c) Hierarchical Feature Decomposition (HFD)}
            \end{overpic}
    \vspace{-8mm} 
    \caption{\textbf{Architectures of our CFD (a), SFE (b), and HFD (c) modules}.
    Our HFD mainly contains of three pairs of CFD and SFE modules.
    For the downsampled low-frequency layer $\mathbf{\hat{L}}_{4}\in\mathbb{R}^{48\times48\times3}$, we first use a $1\times1$ convolution to increases its channel dimension from 3 to 40.
    Then our CFD separates the feature into context-aware feature and residual feature, which are subsequently refined by $1\times1$ convolution followed by a ReLU function and an SFE module, respectively.
    The three hierarchical context-aware feature maps and the residual feature from the third SFE module are summed and fused by a $1\times1$ convolution, with decreased channel dimension from 40 to 8.
    Finally, the fused feature is reshaped into a 3D bilateral grid of affine transformation coefficients $\mathcal{B}\in\mathbb{R}^{16\times16\times72}$.}
    \label{fig:HFD}
    \vspace{-5mm} 
\end{figure*}

\noindent
\textbf{Low-Frequency Layer Correction} is performed by learning pixel-adaptive exposure correction under the bilateral gird learning framework~\cite{xu2021bilateral}.
To learn meaningful bilateral grid of affine coefficients, we propose a parameter-free Context-aware Feature Decomposition (CFD) module and extend it to a hierarchical version for better performance.

\noindent
\textbf{High-Frequency Layers Correction} is implemented by multiplying each layer pixel-wisely with a comfortable mask, predicted by two consecutive $1\times1$ convolutions.

\vspace{2mm}

\noindent
\textbf{Final Reconstruction} is performed by Laplacian reconstruction~\cite{burt1987laplacian} on the exposure-corrected layers of different frequencies to output a well-exposed $\mathbf{O}\in\mathbb{R}^{H\times W\times3}$.
\subsection{Low-Frequency Layer Correction}\label{sec:low-frequency}

The illumination information is mainly in low-frequency~\cite{afifi2021learning}, so we pay more attention to the low-frequency layer $\mathbf{L}_{n}$ for effective exposure correction. Inspired by its success on efficient image processing~\cite{chen2016bilateral,zheng2021ultra,xu2021bilateral}, we employ the bilateral grid learning~\cite{chen2007real} to correct the exposure of low-frequency layer $\mathbf{L}_{n}$. As shown in Figure \ref{fig:MSLT}, our Bilateral Grid Network contains three components: 1) learning the guidance map, 2) estimating the bilateral grid of affine coefficients, and 3) coefficients transformation.

\noindent
\textbf{Learning guidance map}. We propose a Self-modulated Feature Extraction (SFE) module to learn the guidance map $G$ with the same size as $\mathbf{L}_{n}$. As shown in Figure \ref{fig:HFD} (b), the SFE module uses two $1\times1$ convolutions and global average pooling (GAP) to modulate the extracted feature map.

\noindent
\textbf{Estimating bilateral grid of affine coefficients}. We first downsample the low-frequency layer $\mathbf{L}_{n}$ to $\mathbf{\hat{L}}_{n}\in\mathbb{R}^{48\times48\times3}$. The mean and standard deviation (std) of each channel roughly reflect the brightness and contrast, respectively, of that feature map~\cite{IN2016}. Exploiting these information is useful to estimate the bilateral grid of affine coefficients for exposure correction. For this, we propose a parameter-free Context-aware Feature Decomposition (CFD) module to extract the context-aware feature and the residual feature. As shown in Figure \ref{fig:HFD} (a), the context-aware feature is obtained by multiplying the original feature channel-wisely with the sum of mean and std calculated by global average pooling and global std pooling, respectively.
We extend CFD to a Hierarchical Feature Decomposition (HFD) module by cascading three parameter-sharing CFD and SFE modules, as shown in Figure \ref{fig:HFD} (c).
The goal is to learn a 3D bilateral grid of affine coefficients $\mathcal{B}\in\mathbb{R}^{16\times16\times72}$, in which every 12 channels representing a 3$\times$4 affine matrix.
We implement our HFD module by channel-wise $1\times1$ convolutions to perform spatial consistent and pixel-adaptive brightness adjustment. Three $1\times1$ convolutions shared parameters before ReLU, with small parameter amounts and computational costs (Figure \ref{fig:HFD} (c)).

\noindent
\textbf{Coefficients transformation}. With the guidance map $\mathbf{G}\in\mathbb{R}^{\frac{H}{2^{n-1}}\times\frac{W}{2^{n-1}}}$, we upsample the 3D bilateral grid of affine coefficients $\mathcal{B}\in\mathbb{R}^{16\times16\times72}$ back to a 2D bilateral grid of coefficients $\mathbf{B}\in\mathbb{R}^{\frac{H}{2^{n-1}}\times\frac{W}{2^{n-1}}}$ and then correct the low-frequency layer $\mathbf{L}_{n}$ by tri-linear interpolation~\cite{chen2016bilateral}. Each cell of grid $\mathbf{B}$ contains a $3\times4$ matrix for pixel-adaptive affine transformation. At last, the affine transformations in $\mathbf{B}$ will act on the low-frequency layer $\mathbf{L}_{n}$ pixel-by-pixel to obtain the exposure-corrected low-frequency layer $\mathbf{\overline{L}}_{n}$.

\subsection{High-Frequency Layers Correction}
\label{sec:High-Frequency}

With the corrected low-frequency layer, now we correct the high-frequency layers $\{\mathbf{H}_i|i=1,...,n-1\}$ in the order of $i=n-1,...,1$.
The correction is implemented by multiplying each high-frequency layer $\mathbf{H}_i$ with a comfortable mask in an element-wise manner.
Each mask is predicted by a small MLP consisted of two $1\times1$ convolutional layers with a LeakyReLU~\cite{maas2013rectifier} between them.

To correct the high-frequency layer $\mathbf{H}_{n-1}$, we first concatenate it with the upsampled low-frequency layer $\mathbf{L}_{n}$ and the upsampled corrected layer $\mathbf{\overline{L}}_{n}$ along the channel dimension.
Then the concatenated layers are put into the small MLP to predict the mask $\mathbf{M}_{n-1}$.
Since the concatenated layers have nine channels, we set the numbers of input and output channels as nine for the first $1\times1$ convolutional layer in the small MLP, and set those of the second $1\times1$ convolutional layer as nine and three, respectively.
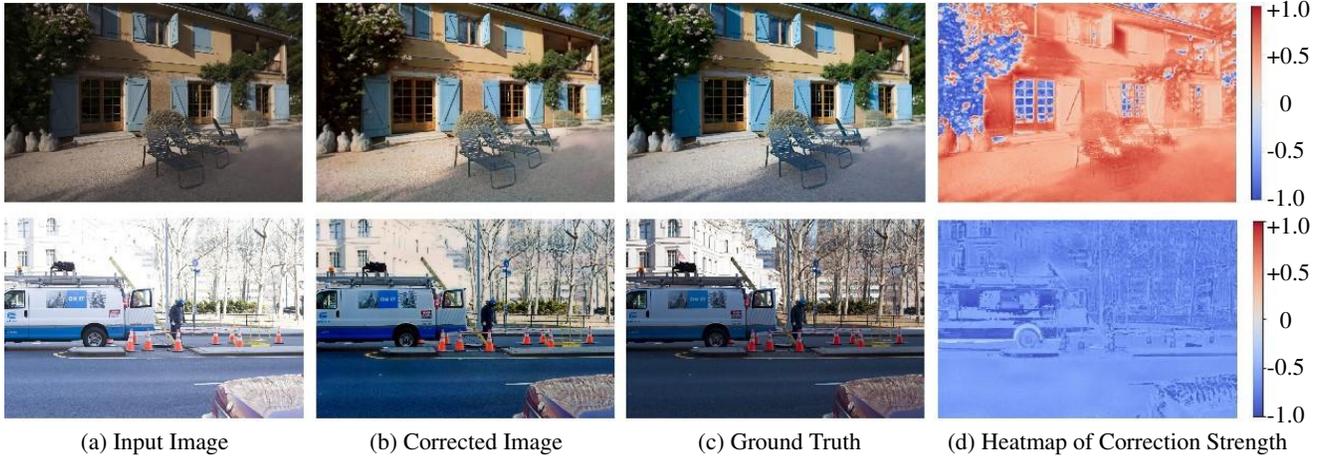
\begin{figure*}[ht]
\vspace{1mm}
    \centering
    \begin{overpic}[width=1\textwidth]{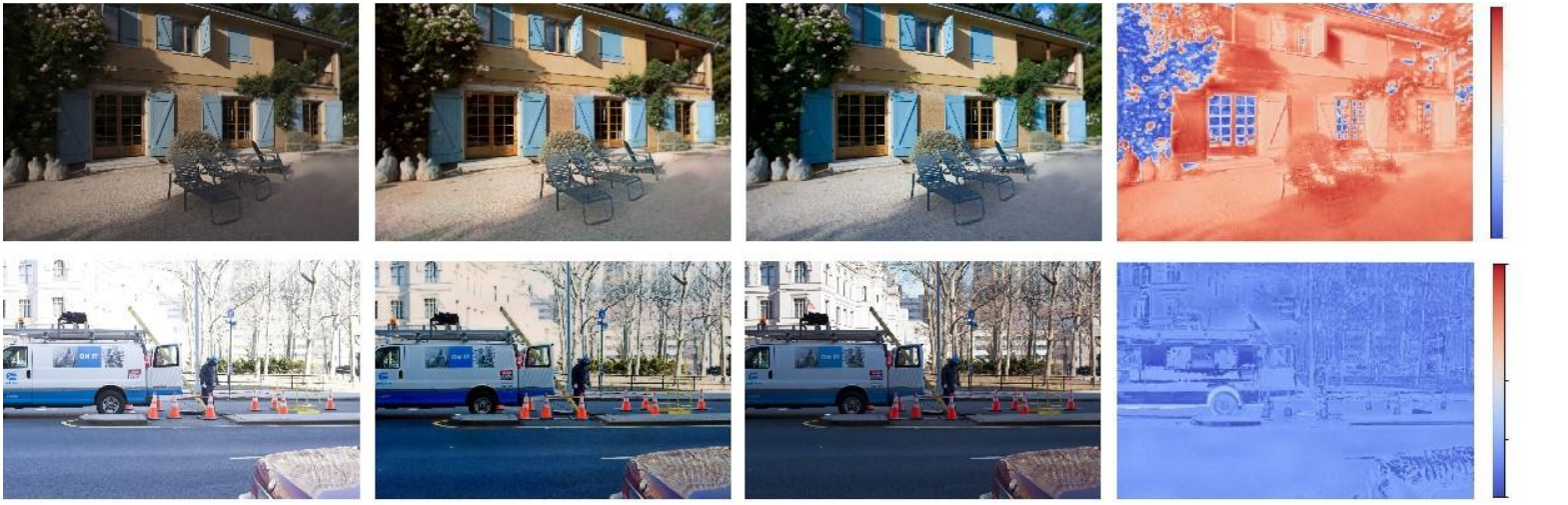}
        \put(6,-2){\small (a) Input Image}
        \put(28,-2){\small (b) Corrected Image }
        \put(53,-2){\small (c) Ground Truth}
        \put(72,-2){\small (d) Heatmap of Correction Strength}
        \put(96.3,31){\small +1.0}
        \put(96.3,27.4){\small +0.5}
        \put(97.3,23.8){\small 0}
        \put(96.3,20.2){\small -0.5}
        \put(96.3,16.6){\small -1.0}
        \put(96.3,14.5){\small +1.0}
        \put(96.3,10.9){\small +0.5}
        \put(97.3,7.3){\small 0}
        \put(96.3,3.7){\small -0.5}
        \put(96.3,0.1){\small -1.0}
    \end{overpic}
    \vspace{-2mm}
    \caption{\textbf{Heatmap of Correction Strength} in our MSLT. (a) the under/over exposed input images. 
    (b) the corrected images by our MSLT.
    (c) the ``ground truth'' images.
    (d) the heatmaps of correction strength described in \S\ref{sec:High-Frequency}.
    The values in $(0,1]$ (or $[-1,0)$) indicate brightness enhancement (or shrinkage).
    Darker color indicates larger absolute values and stronger correction strength in brightness.}

    \label{fig:BG}
    \vspace{-4mm}
\end{figure*}
By element-wisely multiplying high-frequency layer $\mathbf{H}_{n-1}$ with the mask $\mathbf{M}_{n-1}$, we obtain the exposure corrected high-frequency layer $\mathbf{\overline{H}}_{n-1}$.
Besides, the predicted mask $\mathbf{M}_{n-1}$ will be reused as the input of the MLP in the correction of next high-frequency layer for mask prediction.

For $i=n-2,...,1$, we upsample the mask $\mathbf{M}_{i+1}$ output in previous layer into the MLP of current layer to predict a new mask $\mathbf{M}_{i}$.
Unlike the MLP in predicting the mask $\mathbf{M}_{n-1}$, the MLPs for predicting masks $\{\mathbf{M}_{i+1}|i=n-2,...,1\}$ have three input and output channels for both two $1\times1$ convolutional layers.
Similarly, each mask $\mathbf{M}_{i}$ is multiplied with the high-frequency layer $\mathbf{H}_{i}$ element-wisely to output the exposure-corrected high-frequency layer $\mathbf{\overline{H}}_{i}$.
%
Finally, we reconstruct the output image $\mathbf{O}$ from the exposure-corrected low/high-frequency layers $\{\mathbf{\overline{H}}_{1},...,\mathbf{\overline{H}}_{n-1},\mathbf{\overline{L}}_{n}\}$. Here, we set $n=4$ for our MSLT.

To study the effect of exposure correction by our MSLT, we convert the input image $\mathbf{I}$ and output image $\mathbf{O}$ from the sRGB color space to the CIELAB color space.
We denote the lightness channels of $\mathbf{I}$ 
and $\mathbf{O}$ as $\mathbf{I}_L$ and $\mathbf{O}_L$, respectively, and compute their difference residual $\mathbf{R}=\mathbf{O}_L-\mathbf{I}_L$.
Denote $\mathbf{R}_{max}$ as the maximum absolute value of $\mathbf{R}$, \ie, $\mathbf{R}_{max}=\max|\mathbf{R}|$.
The residual $\mathbf{R}$ is normalized into $[-1,1]$ by $\mathbf{R}/\mathbf{R}_{max}$ to represent pixel-wise correction strength, where $(0,1]$ (or $[-1,0)$) indicates brightness enhancement (or shrinkage).
The heatmap of correction strength, as shown in Figure \ref{fig:BG}, exhibits close relationship to the context of input $\mathbf{I}$.
This demonstrates that our MSLT indeed performs pixel-adaptive exposure correction.
\begin{table*}[ht]
\setlength{\abovecaptionskip}{3pt}
\caption{\textbf{Quantitative results of different methods} on the ME dataset~\cite{afifi2021learning}.
We take the correctly exposed images rendered by five experts as the ground truth images, respectively. The best, second best and third best results are highlighted in {\color{red}{\textbf{red}}}, {\color{blue}{\textbf{blue}}} and \textbf{bold}, respectively.
}
\label{table:ME}
\vspace{0.5mm}
\resizebox{\textwidth}{!}{
\begin{tabular}{lllllllllllllllllll}
\hline
\multicolumn{1}{|c|}{\multirow{2}{*}{Method}} & \multicolumn{3}{c|}{Expert A} & \multicolumn{3}{c|}{Expert B}                                                                & \multicolumn{3}{c|}{Expert C}                                                                & \multicolumn{3}{c|}{Expert D}                                                                & \multicolumn{3}{c|}{Expert E}                                                                & \multicolumn{3}{c|}{Average}         
\\ \cline{2-19} 

\multicolumn{1}{|c|}{}                        & \multicolumn{1}{l|}{PSNR$\uparrow$}   & \multicolumn{1}{l|}{SSIM$\uparrow$}         & \multicolumn{1}{l|}{LPIPS$\downarrow$} & \multicolumn{1}{l|}{PSNR$\uparrow$}   & \multicolumn{1}{l|}{SSIM$\uparrow$}         & \multicolumn{1}{l|}{LPIPS$\downarrow$} & \multicolumn{1}{l|}{PSNR$\uparrow$}   & \multicolumn{1}{l|}{SSIM$\uparrow$}         & \multicolumn{1}{l|}{LPIPS$\downarrow$} & \multicolumn{1}{l|}{PSNR$\uparrow$}   & \multicolumn{1}{l|}{SSIM$\uparrow$}         & \multicolumn{1}{l|}{LPIPS$\downarrow$} & \multicolumn{1}{l|}{PSNR$\uparrow$}   & \multicolumn{1}{l|}{SSIM$\uparrow$}         & \multicolumn{1}{l|}{LPIPS} & \multicolumn{1}{l|}{PSNR$\uparrow$}   & \multicolumn{1}{l|}{SSIM$\uparrow$}         & \multicolumn{1}{l|}{LPIPS$\downarrow$} \\ \hline

\multicolumn{1}{|r|}{LPTN~\cite{liang2021high}}
    & \multicolumn{1}{l}{17.50} & \multicolumn{1}{l}{0.746} & \multicolumn{1}{l|}{0.2236}  
    & \multicolumn{1}{l}{18.28} & \multicolumn{1}{l}{0.789} & \multicolumn{1}{l|}{0.2067}  
    & \multicolumn{1}{l}{18.08} & \multicolumn{1}{l}{0.780} & \multicolumn{1}{l|}{0.2121}  
    & \multicolumn{1}{l}{17.70} & \multicolumn{1}{l}{0.770} & \multicolumn{1}{l|}{0.2154}  
    & \multicolumn{1}{l}{17.45} & \multicolumn{1}{l}{0.768} & \multicolumn{1}{l|}{0.2235}  
    & \multicolumn{1}{l}{17.80} & \multicolumn{1}{l}{0.771} & \multicolumn{1}{l|}{0.2519} \\ 

\multicolumn{1}{|r|}{Zero-DCE~\cite{guo2020zero}}          & \multicolumn{1}{l}{12.16} & \multicolumn{1}{l}{0.658} & \multicolumn{1}{l|}{0.3103} &
\multicolumn{1}{l}{13.16} & \multicolumn{1}{l}{0.725} & \multicolumn{1}{l|}{0.2649} & \multicolumn{1}{l}{12.61} & \multicolumn{1}{l}{0.694} & \multicolumn{1}{l|}{0.3022} &\multicolumn{1}{l}{13.47} & \multicolumn{1}{l}{0.720} & \multicolumn{1}{l|}{0.2678} & \multicolumn{1}{l}{14.18} & \multicolumn{1}{l}{0.749} & \multicolumn{1}{l|}{0.2643} & \multicolumn{1}{l}{13.11} & \multicolumn{1}{l}{0.709} & \multicolumn{1}{l|}{0.2819}\\

\multicolumn{1}{|r|}{SCI~\cite{ma2022toward}}          & \multicolumn{1}{l}{16.11} & \multicolumn{1}{l}{0.737} & \multicolumn{1}{l|}{0.2064} & \multicolumn{1}{l}{17.15} & \multicolumn{1}{l}{0.805} & \multicolumn{1}{l|}{0.1725} & \multicolumn{1}{l}{16.36} & \multicolumn{1}{l}{0.764} & \multicolumn{1}{l|}{0.2079} & \multicolumn{1}{l}{16.51} & \multicolumn{1}{l}{0.766} & \multicolumn{1}{l|}{0.1899} & \multicolumn{1}{l}{16.09} & \multicolumn{1}{l}{0.761} & \multicolumn{1}{l|}{0.2125} & \multicolumn{1}{l}{16.44} & \multicolumn{1}{l}{0.767} & \multicolumn{1}{l|}{0.1978} \\ 
\hdashline
\multicolumn{1}{|r|}{MSEC w/o\ adv~\cite{afifi2021learning}}          & \multicolumn{1}{l}{19.16} & \multicolumn{1}{l}{0.796} & \multicolumn{1}{l|}{0.1802} & \multicolumn{1}{l}{20.10} & \multicolumn{1}{l}{0.815} & \multicolumn{1}{l|}{0.1724} & \multicolumn{1}{l}{20.21} & \multicolumn{1}{l}{0.817} & \multicolumn{1}{l|}{0.1805} & \multicolumn{1}{l}{18.98} & \multicolumn{1}{l}{0.796} & \multicolumn{1}{l|}{0.1816} & \multicolumn{1}{l}{18.98} & \multicolumn{1}{l}{0.805} & \multicolumn{1}{l|}{0.1911} & \multicolumn{1}{l}{19.48} & \multicolumn{1}{l}{0.806} & \multicolumn{1}{l|}{0.1812} \\

\multicolumn{1}{|r|}{MSEC  w/  adv~\cite{afifi2021learning}}          & \multicolumn{1}{l}{19.11} & \multicolumn{1}{l}{0.784} & \multicolumn{1}{l|}{0.1861} & \multicolumn{1}{l}{19.96} & \multicolumn{1}{l}{0.813} & \multicolumn{1}{l|}{0.1802} & \multicolumn{1}{l}{20.08} & \multicolumn{1}{l}{0.815} & \multicolumn{1}{l|}{0.1875} & \multicolumn{1}{l}{18.87} & \multicolumn{1}{l}{0.793} & \multicolumn{1}{l|}{0.1901} & \multicolumn{1}{l}{18.86} & \multicolumn{1}{l}{0.803} & \multicolumn{1}{l|}{0.1999} & \multicolumn{1}{l}{19.38} & \multicolumn{1}{l}{0.802} & \multicolumn{1}{l|}{0.1888} \\ 


\multicolumn{1}{|r|}{LCDP~\cite{wang2022local}}          & \multicolumn{1}{l}{{\color{blue}{\textbf{20.59}}}} & \multicolumn{1}{l}{\color{blue}{\textbf{0.814}}} & \multicolumn{1}{l|}{\color{red}{\textbf{0.1540}}} & \multicolumn{1}{l}{21.95} & \multicolumn{1}{l}{0.845} & \multicolumn{1}{l|}{\color{red}{\textbf{0.1399}}} & \multicolumn{1}{l}{\color{blue}{\textbf{22.30}}} & \multicolumn{1}{l}{\color{blue}{\textbf{0.856}}} & \multicolumn{1}{l|}{\color{red}{\textbf{0.1448}}} & \multicolumn{1}{l}{20.22} & \multicolumn{1}{l}{0.825} & \multicolumn{1}{l|}{\color{red}{\textbf{0.1526}}} & \multicolumn{1}{l}{{20.07}} & \multicolumn{1}{l}{0.827} & \multicolumn{1}{l|}{\color{red}{\textbf{0.1617}}} & \multicolumn{1}{l}{{21.02}} & \multicolumn{1}{l}{{\textbf{0.833}}} & \multicolumn{1}{l|}{\color{red}{\textbf{0.1506}}} \\ 
\multicolumn{1}{|r|}{IAT~\cite{90K}}          & \multicolumn{1}{l}{19.63} & \multicolumn{1}{l}{0.780} & \multicolumn{1}{l|}{0.1962} & \multicolumn{1}{l}{21.21} & \multicolumn{1}{l}{0.816} & \multicolumn{1}{l|}{0.1771} & \multicolumn{1}{l}{21.21} & \multicolumn{1}{l}{0.820} & \multicolumn{1}{l|}{0.1828} & \multicolumn{1}{l}{19.58} & \multicolumn{1}{l}{0.805} & \multicolumn{1}{l|}{0.1871} & \multicolumn{1}{l}{19.21} & \multicolumn{1}{l}{0.797} & \multicolumn{1}{l|}{0.1947} & \multicolumn{1}{l}{20.17} & \multicolumn{1}{l}{0.804} & \multicolumn{1}{l|}{0.1876} \\ 
\multicolumn{1}{|r|}{FECNet~\cite{huang2022deep}}          & \multicolumn{1}{l}{\color{red}{\textbf{20.73}}} & \multicolumn{1}{l}{\color{red}{\textbf{0.815}}} & \multicolumn{1}{l|}{0.1861} & \multicolumn{1}{l}{\color{red}{\textbf{22.87}}} & \multicolumn{1}{l}{\color{blue}{\textbf{0.861}}} & \multicolumn{1}{l|}{0.1636} & \multicolumn{1}{l}{\color{red}{\textbf{22.92}}} & \multicolumn{1}{l}{\color{red}{\textbf{0.858}}} & \multicolumn{1}{l|}{0.1700} & \multicolumn{1}{l}{\color{red}{\textbf{20.67}}} & \multicolumn{1}{l}{\color{red}{\textbf{0.835}}} & \multicolumn{1}{l|}{0.1808} & \multicolumn{1}{l}{\textbf{20.22}} & \multicolumn{1}{l}{\textbf{0.829}} & \multicolumn{1}{l|}{0.1913} & \multicolumn{1}{l}{\color{red}{\textbf{21.48}}} & \multicolumn{1}{l}{\color{red}{\textbf{0.839}}} & \multicolumn{1}{l|}{0.1783} \\

\hdashline
\multicolumn{1}{|r|}{Channel-MLP}          & \multicolumn{1}{l}{16.21} & \multicolumn{1}{l}{0.708} & \multicolumn{1}{l|}{0.2577} & \multicolumn{1}{l}{17.48} & \multicolumn{1}{l}{0.784} & \multicolumn{1}{l|}{0.2255} & \multicolumn{1}{l}{16.96} & \multicolumn{1}{l}{0.741} & \multicolumn{1}{l|}{0.2421}  & \multicolumn{1}{l}{16.59} & \multicolumn{1}{l}{0.746} & \multicolumn{1}{l|}{0.2442} & \multicolumn{1}{l}{16.53} & \multicolumn{1}{l}{0.750} & \multicolumn{1}{l|}{0.2481} & \multicolumn{1}{l}{16.75} & \multicolumn{1}{l}{0.746} & \multicolumn{1}{l|}{0.2435} \\ 

\multicolumn{1}{|r|}{MSLT}              & \multicolumn{1}{l}{
{\textbf{20.21}} 
} & 
\multicolumn{1}{l}{{{\textbf{0.805}}}} &
\multicolumn{1}{l|}{{{\textbf{0.1724}}}} & \multicolumn{1}{l}{{22.47}} & \multicolumn{1}{l}{{\color{red}{\textbf{0.864}}}}        & \multicolumn{1}{l|}{0.1460} & \multicolumn{1}{l}{22.03} & \multicolumn{1}{l}{{{\textbf{0.844}}}}        & \multicolumn{1}{l|}{{{\textbf{0.1639}}}} & \multicolumn{1}{l}{{20.33}} & \multicolumn{1}{l}{{\color{blue}{\textbf{0.830}}}}        & \multicolumn{1}{l|}{{{\textbf{0.1637}}}} & \multicolumn{1}{l}{{20.04}} & \multicolumn{1}{l}{{\color{red}{\textbf{0.832}}}}        & \multicolumn{1}{l|}{{{\textbf{0.1758}}}} & \multicolumn{1}{l}{{21.02}} & \multicolumn{1}{l}{{\color{blue}{\textbf{0.835}}}}        & \multicolumn{1}{l|}{{{\textbf{0.1644}}}} \\ 

\multicolumn{1}{|r|}{MSLT+}              & \multicolumn{1}{l}{ {\textbf{20.21}}  
} & \multicolumn{1}{l}{\color{blue}{0.799}}        & \multicolumn{1}{l|}{{\color{blue}{\textbf{0.1677}}}} & \multicolumn{1}{l}{{{\textbf{22.49}}}} & \multicolumn{1}{l}{{0.858}}        & \multicolumn{1}{l|}{{\color{blue}{\textbf{0.1410}}}} & \multicolumn{1}{l}{{{\textbf{22.09}}}} & 
\multicolumn{1}{l}{{0.840}}        & \multicolumn{1}{l|}{{\color{blue}{\textbf{0.1588}}}} & \multicolumn{1}{l}{{\color{blue}{\textbf{20.59}}}} & \multicolumn{1}{l}{{{{\textbf{0.828}}}}}        & \multicolumn{1}{l|}{{\color{blue}{\textbf{0.1585}}}} & \multicolumn{1}{l}{{\color{red}{\textbf{20.53}}}} & \multicolumn{1}{l}{\color{blue}{\textbf{0.830}}}        & \multicolumn{1}{l|}{{\color{blue}{\textbf{0.1687}}}} & \multicolumn{1}{l}{{\color{blue}{\textbf{21.18}}}} & \multicolumn{1}{l}{{0.831}}        & \multicolumn{1}{l|}{{\color{blue}{\textbf{0.1589}}}} \\ 

\multicolumn{1}{|r|}{MSLT++}              & \multicolumn{1}{l}{{20.09}} & \multicolumn{1}{l}{{0.797}}        & \multicolumn{1}{l|}{0.1745} & \multicolumn{1}{l}{{\color{blue}{\textbf{22.55}}}} & 
\multicolumn{1}{l}{\textbf{0.860}}        &
\multicolumn{1}{l|}{{{\textbf{0.1452}}}} &
\multicolumn{1}{l}{{22.07}} & 
\multicolumn{1}{l}{0.838}        & 
\multicolumn{1}{l|}{{{\textbf{0.1639}}}} & \multicolumn{1}{l}{{{{\textbf{20.54}}}}} & \multicolumn{1}{l}{{0.826}}  & \multicolumn{1}{l|}{{0.1640}} & \multicolumn{1}{l}{\color{blue}{\textbf{20.36}}} & \multicolumn{1}{l}{{0.828}}        & \multicolumn{1}{l|}{{0.1762}} & \multicolumn{1}{l}{{{\textbf{21.12}}}} & \multicolumn{1}{l}{{0.830}}        & \multicolumn{1}{l|}{{0.1648}} \\ \hline
\end{tabular}
}
\vspace{-4mm} 
\end{table*}

\begin{table}[h]
\centering
\tiny
 \renewcommand{\arraystretch}{0.65}
 \vspace{0.5mm} 
 \setlength{\abovecaptionskip}{3pt}
 \caption{ \textbf{Quantitative results of different methods} on SICE dataset~\cite{wei2018deep}. The best, second best and third best results are highlighted in {\color{red}{\textbf{red}}}, {\color{blue}{\textbf{blue}}} and \textbf{bold}, respectively.}
 \label{table:SICE}
 \vspace{0.5mm}
 \resizebox{!}{45pt}{%
 \begin{tabular}{|r|lll|}
 \hline
 \renewcommand{\arraystretch}{1.15}
 {\fontsize{4.5}{6}\selectfont Method} & {\fontsize{4.5}{6}\selectfont PSNR$\uparrow$} & {\fontsize{4.5}{6}\selectfont SSIM$\uparrow$} & {\fontsize{4.5}{6}\selectfont LPIPS$\downarrow$}\\ \hline
 
  {\fontsize{4.5}{6}\selectfont LPTN \cite{liang2021high}}&15.46  &0.609  & 0.4150\\
  
{\fontsize{4.5}{6}\selectfont  Zero-DCE \cite{guo2020zero}}& 12.05 & 0.592 & 0.4439\\
 {\fontsize{4.5}{6}\selectfont SCI \cite{ma2022toward}}& 12.85 & 0.569 & 0.3776\\
\hdashline
 {\fontsize{4.5}{6}\selectfont MSEC w/o \ adv \cite{afifi2021learning}}& 17.86 & \textbf{0.664} & \textbf{0.3761}\\
 {\fontsize{4.5}{6}\selectfont MSEC w/ adv \cite{afifi2021learning}}& 17.67 & \textbf{0.664} &0.3875 \\
{\fontsize{4.5}{6}\selectfont  LCDP \cite{wang2022local}}& {18.50} & {{0.609}} & {0.4749} \\
 {\fontsize{4.5}{6}\selectfont IAT \cite{90K}}& \textbf{18.55} & {\color{blue}{\textbf{0.672}}} & {\color{red}{\textbf{0.3325}}}\\ 
  {\fontsize{4.5}{6}\selectfont FECNet \cite{huang2022deep}}& \color{red}\textbf{19.39} & {\color{red}{\textbf{0.691}}} & {0.3939}\\ \hdashline
{\fontsize{4.5}{6}\selectfont  Channel-MLP} & 15.21 & 0.546 &  0.5370 \\
{\fontsize{4.5}{6}\selectfont  MSLT }& 18.22 & 0.661 & {\color{blue}{\textbf{0.3557}}}\\
 {\fontsize{4.5}{6}\selectfont MSLT+ }& 18.32 & 0.642 & 0.3883\\
{\fontsize{4.5}{6}\selectfont MSLT++} & {\color{blue}{\textbf{18.69}}} & 0.653 & 0.3900\\\hline
 \end{tabular}
 }
 \vspace{-3mm}
 \end{table}

\vspace{-5mm}
\subsection{Network Acceleration}
\label{sec:acceleration}
The proposed MSLT network implements Laplacian pyramid decomposition via standard Gaussian kernel~\cite{burt1981fast}, which is not optimized in current deep learning frameworks~\cite{NEURIPS2019pytorch,abadi2016tensorflow}.
To speed up our MSLT, we replace the Gaussian kernel with learnable $3\times3$ convolution kernel, which is highly optimized by the PyTorch framework~\cite{Xin2021learning}.
By introducing $3\times3$ convolutional kernels into our MSLT, we break its fully MLP architecture with more parameters and computational costs.
The speed of our MSLT is clearly improved from 4.34ms to 4.07ms on $1024\times1024$ sRGB images and from 19.27ms to 11.04ms on $3840\times2160$ sRGB images.
We call this variant network as MSLT+.
Through experiments, we also observe that the learnable $3\times3$ convolutional kernels can perform adaptive decomposition for each image to better correct the exposure of different layers.

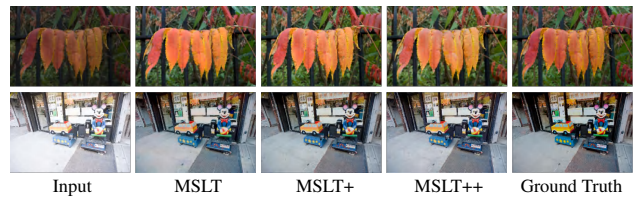
\begin{figure}[t]
    \setlength{\abovecaptionskip}{3pt}
    \centering
    \begin{overpic}[width=0.475\textwidth]{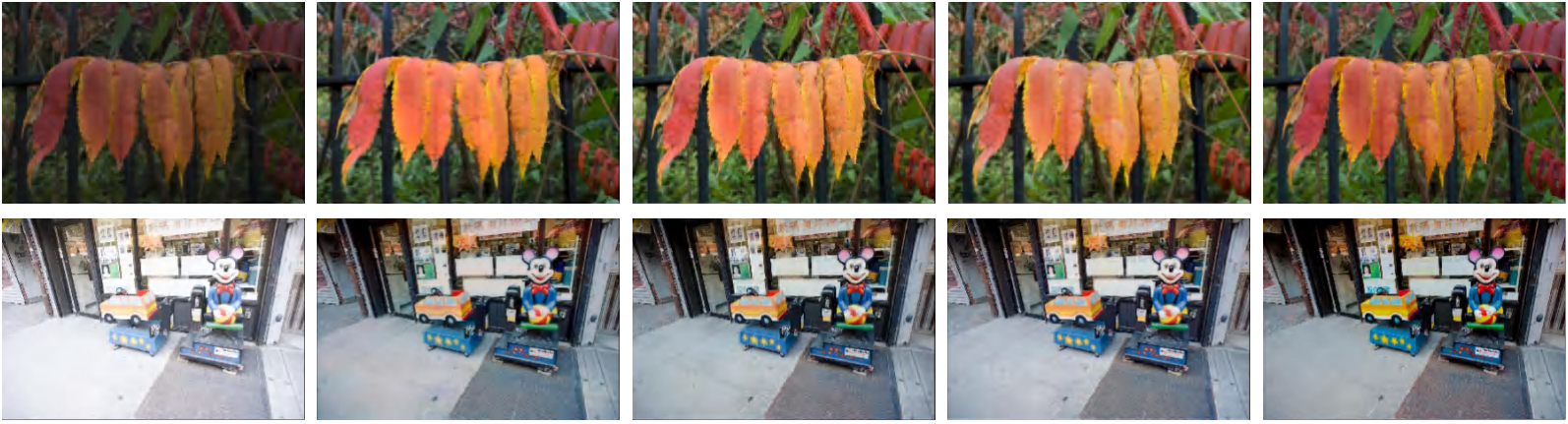}
       \put(7,-3){\scriptsize Input}
       \put(26.5,-3){\scriptsize MSLT}
       \put(46,-3){\scriptsize MSLT+}
       \put(65,-3){\scriptsize MSLT++}
       \put(82,-3){\scriptsize Ground Truth}
    \end{overpic}
    \vspace{0mm}
    \caption{\textbf{Corrected images} by our MSLT, MSLT+ and MSLT++.}
    \label{fig:3.4}
    \vspace{-6mm}   
\end{figure}

Considering that the high-frequency layer $\mathbf{H}_{1}$ is of the largest resolution with the finest information among all layers, it is worth to study whether it is feasible to avoid the correction of this layer for further model acceleration.
In fact, even without correcting $\mathbf{H}_{1}$, the learnable convolution kernels in MSLT+ would still produce adaptive Laplacian pyramid decomposition to compensate the overall exposure correction performance.
To illustrate this point, we remove the mask prediction MLP in correcting the high-frequency layer $\mathbf{H}_{1}$ in MSLT+, and directly using the $\mathbf{H}_{1}$ together with other corrected layers $\{\mathbf{\overline{L}}_{4},\mathbf{\overline{H}}_{3}$,$\mathbf {\overline{H}}_{2}\}$ for final reconstruction.
We call this variant network as MSLT++.
As shown in Figure \ref{fig:3.4}, on two under-exposed and over-exposed images, we observe similar visual quality of the exposure-corrected images by MSLT, MSLT+, and MSLT++.
This indicates that removing the correction of the high-frequency layer $\mathbf{H}_{1}$ potentially influences little our MSLT++ on exposure correction, and brings additional reduction on the computational costs and inference time of MSLT+.
For example, our MSLT++ improves the speed of MSLT+ from 4.07ms to 3.67ms on $1024\times1024$ sRGB images and from 11.04ms to 7.94ms on $3840\times2160$ (4K) sRGB images.

\subsection{Implementation Details}\label{sec:Implementation Details}
Our MSLT networks are optimized by Adam~\cite{kingma2014adam} with $\beta_1$=0.9 and $\beta_2$=0.999, using the mean-square error (MSE) loss function.
The initial learning rate is set as $1\times10^{-3}$ and is decayed to $1\times10^{-7}$ with cosine annealing schedule for every 5 epochs.
%
%
The batch size is 32. For the training set, we randomly crop the images into 512$\times$512 patches.
Here, we have $n=4$ Laplacian pyramid layers, the $64\times64$ low-frequency layers are downsampled to $48\times48$ for learning accurate 3D bilateral grid of affine coefficients.
Our MSLT networks, implemented by PyTorch~\cite{Xin2021learning} and MindSpore~\cite{mindspore}, are trained in 200 epochs on a Titan RTX GPU, which takes about 18 hours.

\section{Experiments}\label{sec:experiment}

\subsection{Dataset and Metric}
\label{sec:Dataset and metric}

\noindent
\textbf{Dataset}.\ We evaluate our MSLT networks on two benchmark datasets: the \textbf{ME} dataset~\cite{afifi2021learning} and the \textbf{SICE} dataset~\cite{cai2018learning}. 

The \textbf{ME} dataset is built upon the MIT-Adobe FiveK dataset~\cite{bychkovsky2011learning}, from which each raw-sRGB image was rendered with five relative exposure values $\{-1.5,-1,0,+1,+1.5\}$ to mimic improperly exposed images. Five expert photographers (A-E) manually retouched the raw-sRGB images to produce the correctly exposed images (``ground truths''). As suggested in~\cite{afifi2021learning}, we use the images retouched by Expert C as the training targets. This dataset contains 17,675 training images, 750 validation images, and 5,905 test images.
%

The \textbf{SICE} dataset is randomly divided into 412, 44, and 100 sequences as train, validation, and test sets respectively.
We set the second and the last second images in each sequence as the under or over exposed inputs, as suggested by~\cite{ENC_2022_CVPR}.
For each image in the training set, we randomly crop 30 patches of size $512\times512$ for  training.

\noindent
\textbf{Evaluation metrics}. We use three evaluation metrics of Peak Signal-to-noise Ratio (PSNR), Structural Similarity Index (SSIM)~\cite{ssim}, and Learned Perceptual Image Patch Similarity (LPIPS)~\cite{zhang2018unreasonable} to measure the distance between the exposure corrected images and the ``ground truths''. For LPIPS, we use the AlexNet~\cite{alexnet} to extract feature maps.

\begin{table}[t]
\vspace{0.5mm}  
\setlength{\abovecaptionskip}{3pt}
\caption{\textbf{Comparison of model size, computational costs, and speed (ms)}. The speed is test on a Titan RTX GPU. MSEC indicates ``MSEC w/o adv''. The best, second best and third best results are highlighted in {\color{red}{\textbf{red}}}, {\color{blue}{\textbf{blue}}} and \textbf{bold}, respectively..}
 \label{table:speed}
 \vspace{0.5mm}
 \resizebox{\linewidth}{!}{%
\begin{tabular}{r|rrrrr|}
\hline
\multicolumn{1}{|c}{\multirow{2}{*}{Method}} & 
\multicolumn{1}{|c}{\multirow{2}{*}{\# Param (K)}}  & 
\multicolumn{2}{c}{FLOPs (G)}& \multicolumn{2}{c|}{Speed (ms)}\\ 
\multicolumn{1}{|}{} &\multicolumn{1}{|c}{}
&\multicolumn{1}{c}{$1024\times1024$} 
&\multicolumn{1}{c}{$3840\times2160$}    
&\multicolumn{1}{c}{$1024\times1024$} & 
\multicolumn{1}{c|}{$3840\times2160$}  
\\\hline
\multicolumn{1}{|r|}{LPTN~\cite{liang2021high}}  &
\multicolumn{1}{r}{616.215} &
\multicolumn{1}{r}{21.55} &
\multicolumn{1}{r}{170.46} &
\multicolumn{1}{r}{6.90} &
\multicolumn{1}{r|}{55.96}
\\ 

\multicolumn{1}{|r|}{Zero-DCE~\cite{guo2020zero}}  &
\multicolumn{1}{r}{79.416}  &
\multicolumn{1}{r}{83.27} &
\multicolumn{1}{r}{658.71} &
\multicolumn{1}{r}{22.98} &
\multicolumn{1}{r|}{197.36}
\\ 

\multicolumn{1}{|r|}{SCI ~\cite{ma2022toward}}  &
\multicolumn{1}{r}{\color{red}{\textbf{0.348}}} &
\multicolumn{1}{r}{0.55} &
\multicolumn{1}{r}{4.38} &
\multicolumn{1}{r}{6.55} &
\multicolumn{1}{r|}{48.37}
\\ \hdashline
\multicolumn{1}{|r|}{MSEC ~\cite{afifi2021learning}}  &
\multicolumn{1}{r}{7015.449} &
\multicolumn{1}{r}{73.35} &
\multicolumn{1}{r}{579.98} &
\multicolumn{1}{r}{240.46} &
\multicolumn{1}{r|}{2250.74}
\\ 


\multicolumn{1}{|r|}{LCDP~\cite{wang2022local}}  &
\multicolumn{1}{r}{281.758} &
\multicolumn{1}{r}{17.33} &
\multicolumn{1}{r}{127.79} &
\multicolumn{1}{r}{507.67} &
\multicolumn{1}{r|}{3305.73}\\

\multicolumn{1}{|r|}{IAT~\cite{90K}}  &
\multicolumn{1}{r}{86.856} &
\multicolumn{1}{r}{22.96} &
\multicolumn{1}{r}{182.59} &
\multicolumn{1}{r}{153.96} &
\multicolumn{1}{r|}{1226.73}\\

\multicolumn{1}{|r|}{FECNet~\cite{huang2022deep}}  &
\multicolumn{1}{r}{151.97} &
\multicolumn{1}{r}{94.61} &
\multicolumn{1}{r}{748.35} &
\multicolumn{1}{r}{139.12} &
\multicolumn{1}{r|}{1277.24}\\

\hdashline
\multicolumn{1}{|r|}{Channel-MLP}  &
\multicolumn{1}{r}{\textbf{7.683}} &
\multicolumn{1}{r}{8.05} &
\multicolumn{1}{r}{63.73} &
\multicolumn{1}{r}{8.69} &
\multicolumn{1}{r|}{66.87}
\\ 
\multicolumn{1}{|r|}{MSLT}  &
\multicolumn{1}{r}{\color{blue}{\textbf{7.594}}} &
\multicolumn{1}{r}{\color{red}{{\textbf{0.08}}}} &
\multicolumn{1}{r}{\color{red}{\textbf{0.42}}} &
\multicolumn{1}{r}{{\textbf{4.34}}} &
\multicolumn{1}{r|}{{\textbf{19.27}}}
\\ 
\multicolumn{1}{|r|}{MSLT+}  &
\multicolumn{1}{r}{8.098} &
\multicolumn{1}{r}{\textbf{0.17}} &
\multicolumn{1}{r}{\textbf{1.10}} &
\multicolumn{1}{r}{\color{blue}{\textbf{4.07}}} &
\multicolumn{1}{r|}{\color{blue}{\textbf{11.04}}}
\\ 
\multicolumn{1}{|r|}{MSLT++}  &
\multicolumn{1}{r}{8.098} &
\multicolumn{1}{r}{\color{blue}{\textbf{0.14}}} &
\multicolumn{1}{r}{\color{blue}{\textbf{0.88}}} &
\multicolumn{1}{r}{\color{red}{\textbf{3.67}}} &
\multicolumn{1}{r|}{\color{red}{\textbf{7.94}}}
\\ \hline
\end{tabular}
}
\vspace{-5mm}
\end{table}

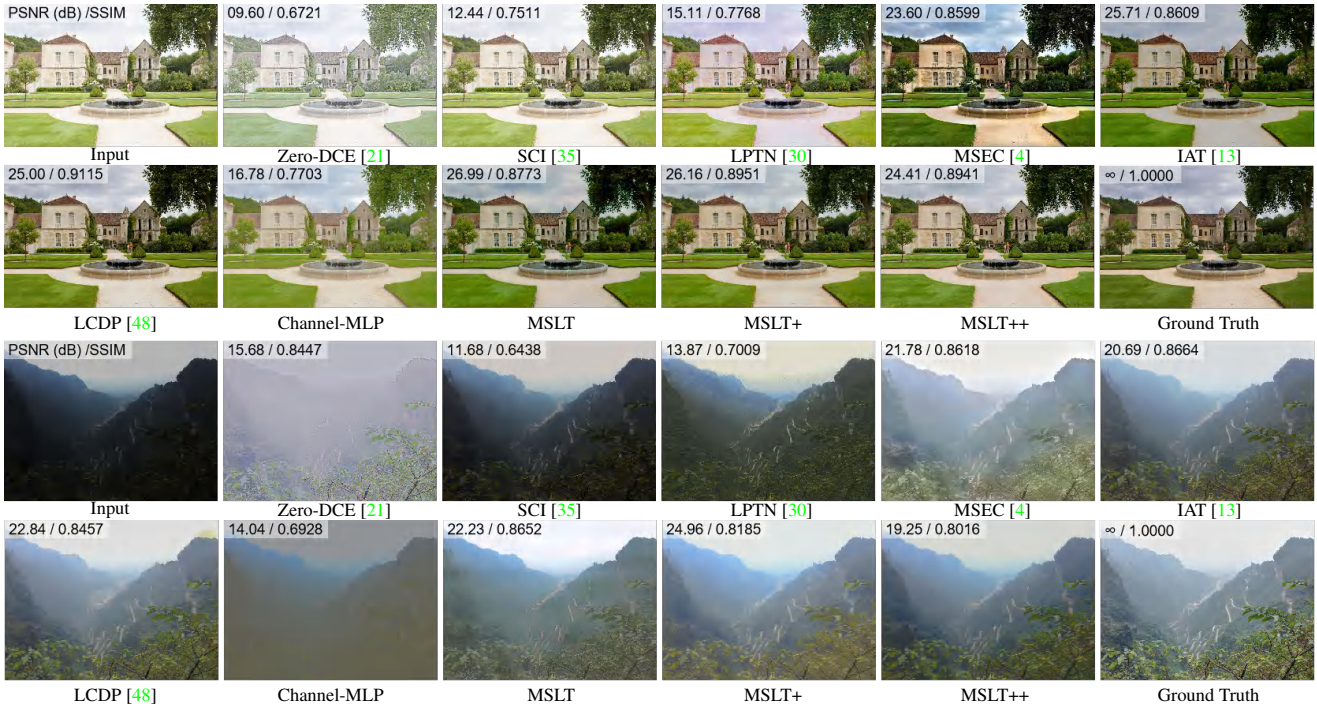
\begin{figure*}[h]
  \vspace{-2mm}
  \centering
  \begin{overpic}[width=1\textwidth]{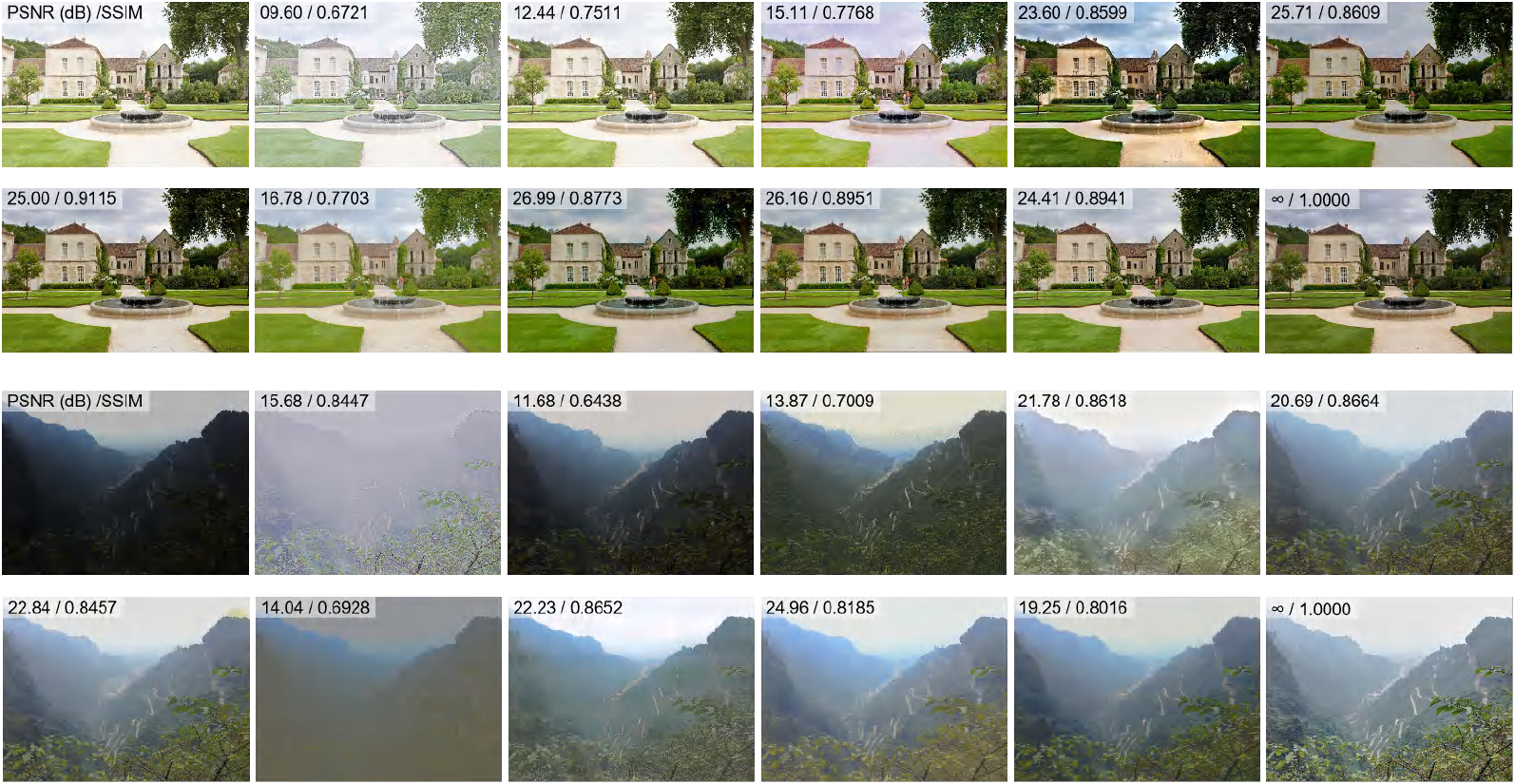}
   \put(6.75,39.8){\scriptsize Input}
   \put(21,39.65){\scriptsize Zero-DCE~\cite{guo2020zero}}
   \put(39.25,39.65){\scriptsize SCI~\cite{ma2022toward}}
   \put(55.5,39.65){\scriptsize LPTN~\cite{liang2021high}}
   \put(72.5,39.65){\scriptsize MSEC~\cite{afifi2021learning}}
   \put(89.5,39.65){\scriptsize IAT~\cite{90K}}

   \put(5.5,26.9){\scriptsize LCDP \cite{wang2022local}}
  \put(21,26.9){\scriptsize Channel-MLP}
  \put(40,26.9){\scriptsize MSLT}
  \put(56.5,26.9){\scriptsize MSLT+}
  \put(73,26.9){\scriptsize MSLT++}
  \put(88,26.9){\scriptsize Ground Truth}
   
   \put(6.75,12.8){\scriptsize Input}
   \put(21,12.65){\scriptsize Zero-DCE~\cite{guo2020zero}}
   \put(39.25,12.65){\scriptsize SCI~\cite{ma2022toward}}
   \put(55.5,12.65){\scriptsize LPTN~\cite{liang2021high}}
   \put(72.5,12.65){\scriptsize MSEC~\cite{afifi2021learning}}
   \put(89.5,12.65){\scriptsize IAT~\cite{90K}}
   
   \put(5.5,-1.45){\scriptsize LCDP \cite{wang2022local}}
   \put(21,-1.45){\scriptsize Channel-MLP}
   \put(40,-1.45){\scriptsize MSLT}
   \put(56.5,-1.45){\scriptsize MSLT+}
   \put(73,-1.45){\scriptsize MSLT++}
   \put(88,-1.45){\scriptsize Ground Truth}
  \end{overpic}
  \vspace{-3mm}
  \caption{\textbf{Visual quality comparison of exposure corrected images by different methods}. 1st and 2nd rows: visual results on one over-exposed image from the ME dataset~\cite{afifi2021learning}. 3rd and 4th rows: visual results on one under-exposed image from the SICE dataset~\cite{cai2018learning}.
  }
  \vspace{-6mm} 
  \label{fig:vis}
 \end{figure*}

\subsection{Comparison Results}\label{sec:Comparison with SOTA}

%
We compare our MSLTs with four exposure correction methods (MSEC~\cite{afifi2021learning}, LCDP~\cite{wang2022local}, FECNet~\cite{huang2022deep} and IAT~\cite{90K}), two enhancement methods (Zero-DCE~\cite{guo2020zero} and SCI~\cite{ma2022toward}), and one image translation method (LPTN~\cite{liang2021high}). To validate the design of our MSLTs with MLPs, we also compare with a plain Channel-MLP with 7,683 parameters (more details are provided in the \textsl{Supplementary File}).

\noindent
\textbf{Objective results}. For the ME and SICE datasets, as shown in Table \ref{table:ME} and Table \ref{table:SICE}, our MSLTs obtain better PSNR, SSIM and LPIPS results than LPTN, Zero-DCE, SCI and Channel-MLP.
%
On ME, our MSLTs achieve better results than MSEC and IAT, and are comparable to LCDP and FECNet.
On SICE, our MSLTs achieve comparable performance with MSECs and a little inferior results to IAT and FECNet.
However, our MSLTs exhibit higher efficiency than all the other comparison methods, as shown in Table~\ref{table:speed}.



\noindent
\textbf{Speed}. In order to be deployed into practical application, the inference speed is put forward high requirements. 
To measure the speed of the models, we randomly generate an ``image'' of size $1024\times 1024\times3$ or 3840$\times$2160$\times$3, repeat the inference test for 100 times, and average the results as the speed of comparison methods.
The speed tests are all run on a Titan RTX GPU.
The results are shown in Table \ref{table:speed}.
One can see that the inference speed of our MSLT++ on a $1024\times1024\times3$ tensor is 3.67 ms, much faster than all the other methods.
On a high-resolution tensor of size $3840\times2160\times3$, our MSLT++ reaches an inference speed of 7.94ms, also faster than the other comparison methods.

\noindent
\textbf{Visual quality}. The ultimate goal of exposure correction task is to restore more realistic images and improve the visual experience of the observer.
Thus, the visual quality of images is also an important factor to consider.
%
In Figure \ref{fig:vis}, we provide the corrected images of ``Manor'' in ME dataset and ``Mountain'' in SICE dataset by the comparison methods, respectively. More visual comparison results can be found in the \textsl{Supplementary File}. On over-exposed ``Manor'' image, one can see that Zero-DCE, SCI, LPTN and Channel-MLP are hardly able to weaken the exposure. Our MSLTs generate better details in clouds, walls and lawns than those of LCDP and IAT. The corrected image by MSEC has too high contrasts to be realistic. On under-exposed ``Mountain'', our MSLTs outperform the others in terms of overall brightness and details of the green leaves.

\subsection{Ablation Study}
\label{sec:ablation}
Here, we provide detailed experiments of our MSLT  on exposure correction to study:
1) the number of Laplacian pyramid layers in our MSLT;
2) how to design the Context-aware Feature Decomposition (CFD) module;
3) the number of CFD modules in our HFD;
4) how to develop the Hierarchical Feature Decomposition (HFD) module in the bilateral grid network;
5) how the correction of high-frequency layers influences our MSLT and MSLT+.
All experiments are performed on the ME dataset~\cite{afifi2021learning}. The images retouched by five experts are respectively considered as the ``ground-truth" images to calculate average PSNR, SSIM and LPIPS values.
We compute FLOPs and speed on a 1024$\times$1024 sRGB image.
The rows with light shadow indicate the results of our MSLT networks on exposure correction.
More results are provided in \textsl{Supplementary File}.

\noindent
\textbf{1) The number of Laplacian pyramid (LP) layers in our MSLT}.
The Laplacian pyramid structure is deployed in our MSLT networks to reduce the computational costs and inference time (speed). As shown in Table \ref{table:LP}, generally, the Laplacian pyramid with more layers produces smaller low-frequency layer. Since the main costs are paid to this layer, our MSLT will be faster.
However, when the number of LP layers is 5, the low-frequency layer is small, which degrades our MSLT network. Besides, the decomposition of 5 LP layers offsets the overall acceleration, and slow down our MSLT for exposure correction. By considering both the performance and inference speed of our MSLT, we set $n=4$ for the LP decomposition in our MSLT networks.

\begin{table}[t]
\vspace{-0mm} 
\setlength{\abovecaptionskip}{3pt}
\caption{\textbf{Results of exposure correction by our MSLT with different number ($n$) of Laplacian pyramid levels}. ``w/o\ LP'' means we do not use Laplacian pyramid.}
\vspace{0.5mm}
\label{table:LP}
\resizebox{\linewidth}{!}{%
\begin{tabular}{c|cccrrr}
\Xhline{1pt}
\rowcolor[rgb]{ .9,  .9,  .9}
LP Layers & PSNR $\uparrow$ & SSIM $\uparrow$ & LPIPS $\downarrow$ & \# Param & FLOPs (M) & Speed (ms)
\\ \hline
w/o\ LP & 21.06 & 0.830 & 0.1615 & 7,448& 303.85 & 6.50  \\
2      & 20.98 & 0.835 & 0.1631 & 7,594 & 237.79 &4.91  \\
3      & 20.92 & 0.828 & 0.1643 & 7,594 & 114.32 &4.41  \\
\rowcolor[rgb]{ .95,  .95,  .95}
4      & 21.02 & 0.835 & 0.1644 & 7,594 & 83.45 & 4.34  \\
5      & 20.55 & 0.825 & 0.1646 & 7,594&  75.73& 4.66 \\
\hline
\end{tabular}%
}
\vspace{-6mm} 
\end{table}

\noindent
\textbf{2) How to design the Context-aware Feature Decomposition (CFD) module?}
In our CFD, we use the mean and standard deviation of each channel to learn the context-aware feature. To demonstrate its effect, we replace this part with Instance Normalization (IN)~\cite{ulyanov2016instance} or Channel Attention (CA)~\cite{hu2018squeeze}, and remain the rest of our MSLT. As shown in Table \ref{table:caam}, our CFD achieves highest PSNR and LPIPS among the three methods and it has comparable SSIM with the ``IN'' version. This shows that the method using mean and standard deviation information of each channel does work.

\begin{table}[t]
\vspace{-0mm} 
\setlength{\abovecaptionskip}{3pt}%
\caption{ \textbf{Results of our MSLT with different variants of CFD module in our HFD}. ``CFD": Context-aware Feature Decomposition. ``IN": Instance Normalization~\cite{ulyanov2016instance} with feature decomposition. ``CA": Channel Attention~\cite{hu2018squeeze} with feature decomposition. }
\vspace{0.5mm}
\resizebox{\linewidth}{!}{%
\label{table:caam}
\begin{tabular}{c|cccrrr}
\Xhline{1pt}
\rowcolor[rgb]{ .9,  .9,  .9}
Variant & PSNR $\uparrow$ & SSIM $\uparrow$ & LPIPS $\downarrow$ & \# Param & FLOPs (M) & Speed (ms) \\ \hline
IN &  20.82 & 0.831 & 0.1652 & 7,684 & 83.45 & 4.28 \\ 
CA & 20.60 & 0.829 & 0.1701 & 7,912 &83.45 & 4.22 \\ 
\rowcolor[rgb]{ .95,  .95,  .95}
CFD  & 21.02 & 0.835 & 0.1644 & 7,594 & 83.45 & 4.34  \\ \hline
\end{tabular}%
}
\vspace{-2mm} 
\end{table}

\noindent
\textbf{3) The number of CFD modules in our HFD.}
To better learn bilateral grid of affine coefficients, we extend Context-aware Feature Decomposition (CFD) module to a hierarchical structure. As a comparison, we set different number of CFD modules as the composition of Hierarchical Feature Decomposition (HFD). From Table \ref{table:blocks}, it can be found that when the number of CFD modules of HFD increases from 1 to 5, the performance of our MSLT improves and then decreases, reaching the best results with three CFDs. This demonstrates that the power of context transformation is enhanced by multiple modules. However, it is unnecessary to use too many CFD modules to extract redundant features. Therefore, we use three CFD modules in our HFD module.

\begin{table}[t]
\vspace{-2mm} 
\setlength{\abovecaptionskip}{3pt}
\caption{ \textbf{Results of our MSLT with different number of CFD modules} in the proposed HFD module.}
\vspace{0.5mm}
\label{table:blocks}
\resizebox{\linewidth}{!}{%
\begin{tabular}{c|cccrrr}
\Xhline{1pt}
\rowcolor[rgb]{ .9,  .9,  .9}
\# CFD & PSNR $\uparrow$ & SSIM $\uparrow$ & LPIPS $\downarrow$ & \# Param & FLOPs (M) & Speed (ms) \\ \hline
1 & 20.31 & 0.824 & 0.1845 & 7,594 & 60.59 &3.54 \\
2 & 20.50 & 0.826 & 0.1818 & 7,594 & 72.02 &3.82 \\
\rowcolor[rgb]{ .95,  .95,  .95}
3 & 21.02 & 0.835 & 0.1644 & 7,594 & 83.45 &4.34 \\
4 & 20.73 & 0.832 & 0.1699 & 7,594 & 94.88 & 4.56 \\
5 & 20.63 & 0.827 & 0.1714 & 7,594 & 106.31 & 4.91 \\ \hline

\end{tabular}%
}
\vspace{-6mm} 
\end{table}

\noindent
\textbf{4) How to develop the Hierarchical Feature Decomposition (HFD) module in the bilateral grid network}?
To answer this question, we apply a variety of networks with comparable parameters with our HFD module to conduct experiments. For ease of presentation, we denote the network consisting of multiple $1\times1$ convolutional layers and ReLU activation layers as ``Conv-1''. Similarly, when only using $3\times3$ convolutions, the network is denoted as ``Conv-3''. More details are provided in the \textsl{Supplementary File}. As shown in Table \ref{table:FE}, although ``Conv-1'' and ``Conv-3'' also achieve fast speed, our MSLT with HFD achieves better quantitative results in terms of PSNR, SSIM and LPIPS. This shows that our HFD module well estimates the 3D bilateral grid of affine coefficients for exposure correction.

\noindent
\textbf{5) How the correction of high-frequency layers influences our MSLT and MSLT+?} To this end, for both MSLT and MSLT+, we use partial instead of all corrected high-frequency layers for LP reconstruction. Specifically, our experimental setting could be seen in Table \ref{table:acc}. The $\mathbf{\overline{H}}_{i}$ means that we use the corrected high-frequency layer for LP reconstruction. These high-frequency layers are used for LP reconstruction with  $\mathbf {\overline{L}}_{4}$. Similarly, the $\mathbf{H}_{i}$ means we directly use the unprocessed high-frequency layer for LP reconstruction.  As shown in Table \ref{table:acc}, from $\mathbf {\overline{H}}_{3}$+$\mathbf{\overline{H}}_{2}$+$\mathbf {\overline{H}}_{1}$ to $\mathbf {\overline{H}}_{3}$+$\mathbf {\overline{H}}_{2}$+$\mathbf{H}_{1}$, we clearly reduce the FLOPs and inference time (speed) of our MSLT and MSLT+, with little influence on the objective metrics. In our MSLT+, $\mathbf{H}_{1}$ is generated by learnable convolutions, which can partly compensate for the effect of not processing $\mathbf{H}_{1}$. This is why our acceleration strategy has little impact on the objective results of MSLT+. All these results show that our acceleration strategy applied on MSLT+ influences little on the objective metrics, but can clearly reduce the computational costs and inference speed.
\vspace{-5mm}

\begin{table}[t]
\vspace{0mm}  
\setlength{\abovecaptionskip}{3pt}
 \caption{\textbf{Results of our MSLT with different variants of HFD module} in the developed Bilateral Grid Network. ``Conv-1" (or ``Conv-3"): the network consisting of multiple $1\times1$ (or $3\times3$) convolutional layers and ReLU activation function. ``HFD": our Hierarchical Feature Decomposition module.}
\vspace{0.5mm}
\label{table:FE}
\resizebox{\linewidth}{!}{%
\begin{tabular}{c|cccrrr}
\Xhline{1pt}
\rowcolor[rgb]{ .9,  .9,  .9}
Variant & PSNR $\uparrow$ & SSIM $\uparrow$ & LPIPS $\downarrow$ & \# Param & FLOPs (M) & Speed (ms)\\ \hline
``Conv-1" & 19.31 & 0.810 & 0.2103 & 7,676 & 64.47 &3.58 \\
``Conv-3" & 19.10 & 0.795 & 0.2167 & 8,410 &65.54 & 3.70 \\
\rowcolor[rgb]{ .95,  .95,  .95}
HFD & 21.02 & 0.835 & 0.1644 & 7,594 & 83.45 &4.34 \\ \hline
\end{tabular}%
}
\vspace{0mm} 
\end{table}
\begin{table}[t]
\vspace{-2mm}
\setlength{\abovecaptionskip}{4pt}
\caption{\textbf{Results of our MSLT and MSLT+ with some high-frequency layers in Laplacian pyramid unprocessed by MSLT/MSLT+}. ``$\rm{{H}}_{i}$": the unprocessed high-frequency layer. ``$\rm{\overline{H}}_{i}$": the exposure-corrected high-frequency layer.}
\vspace{0.5mm}
\label{table:acc}
\resizebox{\linewidth}{!}{%
\renewcommand{\arraystretch}{1.15}
\begin{tabular}{c|c|cccrrr} \hline
\rowcolor[rgb]{ .9,  .9,  .9}
Model & Layers & PSNR $\uparrow$ & SSIM$\uparrow$ & LPIPS $\downarrow$ & \# Param & FLOPs (M) &  Speed (ms) \\ \hline
\multirow{4}{1cm}{MSLT} &

$\rm{\overline{H}}_{3}$+$\rm {\overline{H}}_{2}$+$\rm {\overline{H}}_{1}$ & 21.02 & 0.835 & 0.1644 & 7,594 & 83.45 &4.34 \\
&
$\rm {\overline{H}}_{3}$+$\rm {\overline{H}}_{2}$+$\rm {H_1}$ & 20.82 & 0.831 & 0.1704 & 7,594 &55.14 & 3.97 \\
&
$\rm {\overline{H}}_{3}$+$\rm {H_2}$+$\rm {H_1}$ &  20.60 & 0.818 & 0.1841 & 7,568 & 48.06 & 3.72 \\
&
$\rm {H_3}$+$\rm {H_2}$+$\rm {H_1}$ &  20.46 & 0.820 & 0.2004& 7,448 & 39.61 & 3.60 \\  \hline
\multirow{4}{1cm}{MSLT+} &

$\rm {\overline{H}}_{3}$+$\rm {\overline{H}}_{2}$+$\rm {\overline{H}}_{1}$ & 21.18 & 0.831 & 0.1589 & 8,098 & 170.15 & 4.07 \\
&
$\rm {\overline{H}}_{3}$+$\rm {\overline{H}}_{2}$+$\rm {H_1}$ & 21.12 & 0.830 & 0.1648 & 8,098 & 141.84 & 3.67  \\
&
$\rm {\overline{H}}_{3}$+$\rm {H_2}$+$\rm {H_1}$ & 21.15 & 0.827 & 0.1723 & 8,072 & 134.77 & 3.59 \\
&
$\rm {H_3}$+$\rm {H_2}$+$\rm {H_1}$ & 20.57 & 0.817 & 0.1806 & 7,952 & 126.31 &3.36 \\ \hline
\end{tabular}%
}
\vspace{-6mm}
\end{table}


\section{Conclusion}
\label{sec:conclusion}

In this paper, we proposed a light-weight and efficient Multi-Scale Linear Transformation (MSLT) network for exposure correction.
The proposed MSLT sequentially corrects the exposures of multi-scale low/high-frequency layers decomposed by Laplacian pyramid technique.
For the low-frequency layer, we developed a bilateral grid network to learn context-aware affine transformation for pixel-adaptive correction.
The high-frequency layers are multiplied in an element-wise manner by comfortable masks learned by channel-wise MLPs.
We also accelerated our MSLT by learnable multi-scale decomposition and removing the correction of the largest high-frequency layer.
The resulting MSLT++ network has 8,098 parameters, and can process a 4K-resolution image at a 125 FPS speed with only 0.88G FLOPs.
Experiments on two benchmarks demonstrated that, our MSLT networks are very efficient and exhibit promising exposure correction performance.

\noindent
\textbf{Acknowledgements}. Jun Xu is partially sponsored by the National Natural Science Foundation of China (No. 62002176, 62176068, and 62171309), CAAI-Huawei MindSpore Open Fund, and the Open Research Fund (No. B10120210117-OF03) from the Guangdong Provincial Key Laboratory of Big Data Computing, The Chinese University of Hong Kong, Shenzhen.

 {\small
\bibliographystyle{ieee_fullname}
\bibliography{egbib}
}

\clearpage
\section{Appendix}
\label{sec:Content}
In this supplemental file, we provide more details of our Multi-Scale Linear Transformation (MSLT) networks presented in the main paper. Specifically, we provide
\begin{itemize}
    \item the detailed implementation of Laplacian Pyramid (LP) decomposition and reconstruction in \S~\ref{sec:LP}.

    \item the channel dimension of the features in our SFE module in \S~\ref{sec:SFE}.

    \item the details of coefficient transformation in our bilateral grid network in \S~\ref{sec:BGN}.
    
    \item more details of high-frequency layers correction in \S~\ref{sec:high}.

    \item the architecture of the Channel-MLP network in our main paper in \S~\ref{sec:CMLP}.
    
    \item more ablation studies in \S~\ref{sec:Ablation}.  
    
    \item more visual comparisons of our MSLTs with the other comparison methods on the ME~\cite{afifi2021learning} and SICE datasets~\cite{cai2018learning} in  \S~\ref{sec:vis1}.
    
    \item the visual comparisons in ablation studies in \S~\ref{sec:vis2}.
    
    \item the societal impact in \S~\ref{sec:impact}.

\end{itemize}

\subsection{Detailed implementation of Laplacian Pyramid (LP) decomposition and reconstruction}
\label{sec:LP}

In our MSLT, we deploy the conventional Gaussian kernel for Laplacian Pyramid (LP)~\cite{burt1987laplacian,liang2021high,denton2015deep,lai2017deep} decomposition and reconstruction. In decomposition, we first use a fixed $5\times5$ Gaussian kernel (Eqn. \ref{eq:gauss}) to perform convolution on the input image $\mathbf{I}\in\mathbb{R}^{H\times W\times3}$ with stride = 2, padding = 2 to obtain $\mathbf{G}_1$. Then, we perform the same convolution operation on $\mathbf{G}_i$ (${i=1,...,n-1}$, note that $n=4$ in our MSLTs) to generate $\mathbf{G}_{i+1}$. After getting Gaussian pyramid sequence $\{ {\mathbf{G}_i\in\mathbb{R}^{\frac{H}{2^{i-1}}\times\frac{W}{2^{i-1}}\times3}|i=1,...,n}\}$, we upsample the Gaussian pyramid $\mathbf{G}_{i+1}$ (${i=1,...,n-1}$) by inserting comfortable all-zero vectors between every two rows and between every two columns, which is convolved with the Gaussian kernel (Eqn. \ref{eq:gauss}) and then subtracted from $\mathbf{G}_i$ to obtain the high-frequency layer $\mathbf{H_i}$ of Laplacian pyramids.
For ${i=n}$, we directly treat $\mathbf{G}_n$ as the low-frequency layer $\mathbf{L_n}$.
In this way, we obtain the Laplacian pyramids of $\{{\mathbf{H}_i|i=1,...,n-1}\}$ and $\mathbf{L_n}$. In reconstruction, for each layer in the processed Laplacian pyramids, we use the same upsample method used in the decomposition and then add the results to the higher layer. Finally, we obtain the reconstructed image $\mathbf{O}\in\mathbb{R}^{H\times W\times3}$. 
\begin{equation}
\text{Gaussian kernel} = \frac{1}{256}\times \begin{bmatrix}
   1 & 4 & 6 & 4 & 1 \\
   4 & 16 & 24 & 16 & 4 \\
   6 & 24 & 36 & 24 & 6 \\
   4 & 16 & 24 & 16 & 4 \\
   1 & 4 & 6 & 4 & 1 \\
  \end{bmatrix}
  \label{eq:gauss}
\end{equation}

In our MSLT+ and MSLT++, we introduce learnable $3\times3$ convolutions with stride = 2 for downsampling in the Laplacian pyramid decomposition, and 
$3\times3$ convolutions with stride = 1 followed by bi-linear interpolation for upsampling in the Laplacian pyramid reconstruction.

\subsection{Channel dimension of the features in our SFE module}
\label{sec:SFE}
\begin{figure}[h]
    
    \vspace{4mm}
    \centering
    \begin{overpic}[width=0.47\textwidth]{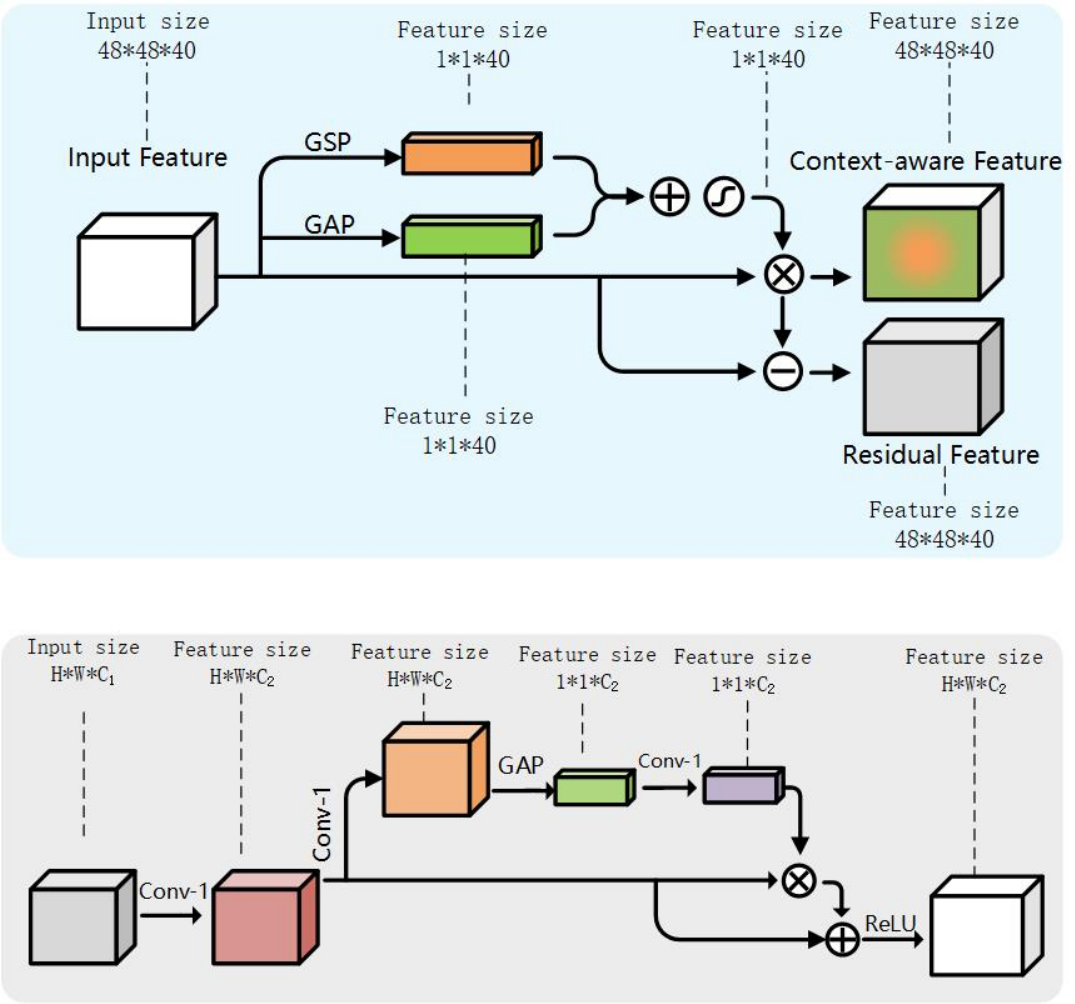}
        \put(9,94.5){(a) Context-aware Feature Decomposition (CFD)}
        \put(12,35.5){(b) Self-modulated Feature Extraction (SFE)}
        \end{overpic}
\vspace{0mm}
    \caption{\textbf{The detailed structure of our CFD module (a) and our SFE module (b)}. In our MSLTs, the CFD module receives a fixed feature input size of $48\times48\times40$ in our HFD. But the input and output of SFE module in predicting the guidance map is different from that in HFD module. See \S~\ref{sec:SFE} for details.}
    \label{fig:CFDSFE}
    \vspace{-6mm}
    
\end{figure}

For our Self-modulated Feature Extraction (SFE) module, as shown in Figure~\ref{fig:CFDSFE} (b), we describe the specific numbers of input channels and output channels for the SFE module, which is used in both predicting the guidance map $\mathbf{G}$ in our bilateral grid network and feature extraction in our Hierarchical Feature Decomposition (HFD) module, as shown in Figures 2 and 3 in the main paper. For the guidance map prediction, numbers of channel $\mathbf{C}_1$ and $\mathbf{C}_2$ are 3 and 8, respectively. In order to generate a gray-scale guidance map, we additionally take a $1\times1$ convolution from 8 channels to 1 channel at the end of SFE. For the feature extraction in our HFD, both $\mathbf{C}_1$ and $\mathbf{C}_2$ are equal to 40.

\subsection{Details of coefficient transformation in our bilateral grid network}
\label{sec:BGN}
Here, we elaborate on the coefficient transformation in the bilateral grid network our MSLT. We use the 3D bilateral grid of affine transformation coefficients $\mathcal{B}\in\mathbb{R}^{16\times16\times72}$ and the guidance map $\mathbf{G}\in\mathbb{R}^{\frac{H}{2^{n-1}}\times\frac{W}{2^{n-1}}}$ for slicing~\cite{chen2016bilateral}.
We compute a 2D grid of coefficients $\mathbf{B}\in\mathbb{R}^{\frac{H}{2^{n-1}}\times\frac{W}{2^{n-1}}}$ using $\mathcal{B}$ and pixel locations from grid $\mathbf{G}$ by tri-linear interpolation~\cite{chen2016bilateral}:
\begin{equation}
\mathbf{B}[x,y] = \sum\limits_{i,j,k} {\tau ({g_h}x - i)} \tau ({g_w}y - j)\tau (d \cdot \mathbf{G}[x,y] - k)\mathcal{B}[i,j,k],
\end{equation}
where $\tau( \cdot ) = max(1-|\cdot|,0)$ is the linear interpolation kernel, ${g_h}$ and ${g_w}$ are the spacial shape of grid $\mathcal{B}$. We fix both ${g_h}$ and ${g_w}$ to 16 and the depth of $\mathcal{B}$ to d = 6.
Each cell of grid $\mathbf{B}$ contains 12 channels. 
For each pixel of the low-frequency layer $\mathbf{L}_n$, we multiply the three RGB values with the corresponding values of the 1st to the 3rd channels of the corresponding pixel in grid $\mathbf{B}$ and add them together, plus the fourth channel value as a bias to get corrected R channel value of the pixel. Similarly, the G and B channels of this pixel are corrected. More details about the bilateral grid learning based transformation scheme can be found in~\cite{chen2016bilateral}.

\subsection{More details of high-frequency layers correction}
\label{sec:high}
For the processing of the high-frequency layers, we deploy a small MLP consisted of two $1\times1$ convolutional layers with a LeakyReLU~\cite{maas2013rectifier} between them.
For high-frequency layer $\mathbf{H}_{n-1}$, when predicting the mask $\mathbf{M}_{n-1}$, the input is a 9-channel feature map concatenated by $\mathbf{H}_{n-1}$, the upsampled low-frequency layer $\mathbf{L}_{n}$ and the upsampled corrected low-frequency layer $\mathbf{\overline{L}}_{n}$ along the channel dimension. So we set the channel numbers of the input and output to the first $1\times1$ convolutional layer as both 9. We set the channel numbers of the input and output to the second $1\times1$ convolutional layer   as 9 and 3, respectively.
For each other high-frequency layer $\mathbf{H}_{i}$ ($i=n-2,...,1$), we set the channel numbers of the input and output to both $1\times1$ convolutional layers as 3 to predict the mask $\mathbf{M}_i$.

Additionally, in our MSLT++ network, we directly use the high-frequency layer $\mathbf{H}_{1}$ for Laplacian pyramid reconstruction rather than that processed by the high-frequency layer correction to accelerate the inference speed. The specific structure of MSLT++ is shown in Figure \ref{fig:MSLT++}.

\subsection{Architecture of the Channel-MLP network in our main paper}
\label{sec:CMLP}
\begin{figure}[h]
    \vspace{-2mm}
    \begin{center}
\includegraphics[width=8cm]{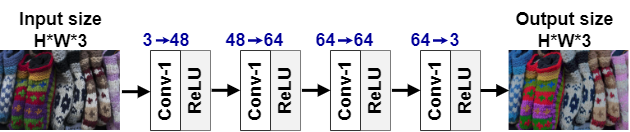}
\vspace{0mm}
    \caption{\textbf{Architecture of the comparison Channel-MLP}. The numbers on the ``Conv-1'' box represent the number of input and output channels of the convolution, respectively.}
    \label{fig:CMLP}
    \end{center}
    \vspace{-6mm}
    
\end{figure}

\begin{figure*}[h]
    \vspace{-4mm}
    \begin{center}
\includegraphics[width=1\textwidth]{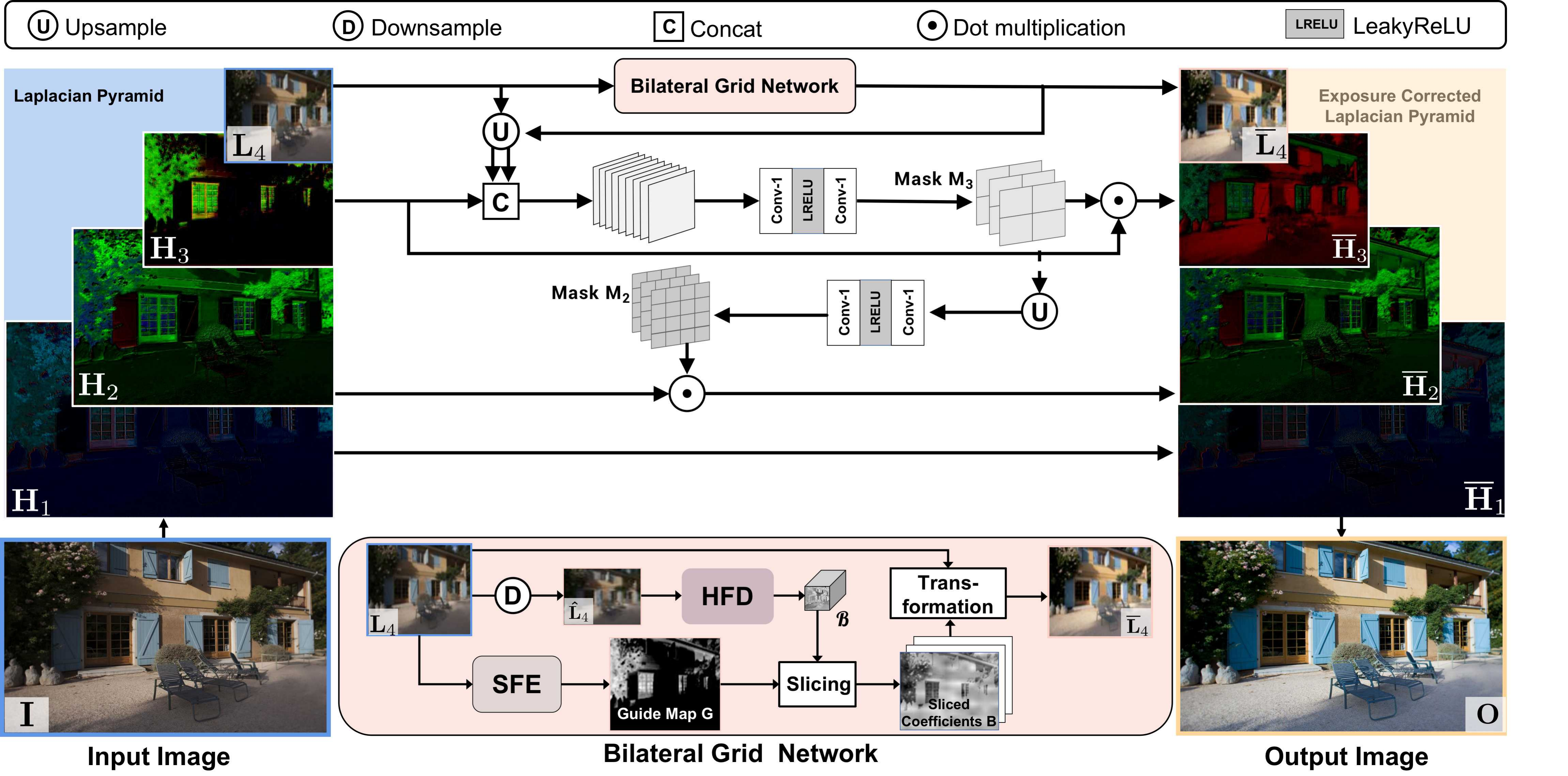}
\vspace{-3mm}
    \caption{\textbf{Overview of our MSLT++ network}. Based on MSLT+ network, we remove the mask prediction MLP in correcting the high-frequency layer $\mathbf{H}_{1}$ in MSLT+, and directly using the $\mathbf{H}_{1}$ together with other corrected layers $\{\mathbf{\overline{L}}_{4},\mathbf{\overline{H}}_{3},\mathbf {\overline{H}}_{2}\}$ for final 
     LP reconstruction.}
    \label{fig:MSLT++}
    \end{center}
    \vspace{-6mm}
    
\end{figure*}

To reduce the parameter amount and computational costs, we employ channel-wise MLP widely in our MSLTs. As a comparison to MLPs, we design a plain Channel-MLP network with 7,683 parameters to perform exposure correction in the Tables 1-3 and Figure 6 of the main paper. As shown in Figure~\ref{fig:CMLP}, the plain Channel-MLP network contains four sequential $1\times1$ convolutional layers, each of which followed by a ReLU activation layer.

\subsection{More Ablation Studies}
\label{sec:Ablation}
In this section, we provide more experimental results to study:
1) how parameter sharing in high-frequency layers correction influences the performance of our MSLT?
2) how the GAP and GSP influence our CFD module?
3) how to design the use of SFE modules in our HFD module?
4) the effect of feature separation order in our CFD module.

\noindent
\textbf{1) How parameter sharing in high-frequency layers correction influences the performance of our MSLT?} In high-frequency layers correction, we deploy small MLPs consisted of two $1\times1$ convolutional layers with a LeakyReLU~\cite{maas2013rectifier} between them to predict Mask $\{{\mathbf{M}_i|i=1,...,n-1} \}$. As described in \S~\ref{sec:high}, the $1\times1$ convolutions used to predict Mask $\{{\mathbf{M}_i|i=1,...,n-2} \}$ has 3 input and output channels. Therefore, we design a comparison experiment of whether small MLPs used in different high-frequency layers correction share parameters. As shown in Table~\ref{table:1}, whether the small MLPs in high-frequency layers correction share parameters has little effect on the performance of our MSLT. For a lower number of parameters, we choose sharing parameters in our MSLT.
\begin{table}[h]
\setlength{\abovecaptionskip}{3pt}
 \caption{\textbf{Results of the high-frequency layers correction of our MSLT with the parameters of $1\times1$ convolutions shared or not.} ``not shared'' means we deploy independent convolutions between each high-frequency layer. ``shared'' means small MLPs in different high-frequency layers share convolution parameters.}
\vspace{0.3mm}
\label{table:1}
\resizebox{\linewidth}{!}{%
\begin{tabular}{c|cccrrr}
\Xhline{1pt}
\rowcolor[rgb]{ .9,  .9,  .9}
Method & PSNR $\uparrow$ & SSIM $\uparrow$ & LPIPS $\downarrow$ & \# Param. & FLOPs (M) & Speed (ms)\\ \hline
not shared & 20.87 & 0.832 & 0.1670 & 7,618 & 83.45 &4.24 \\
\rowcolor[rgb]{ .95,  .95,  .95}
shared & 21.02 & 0.835 & 0.1644 & 7,594 & 83.45 & 4.34  \\ \hline
\end{tabular}%
}
\vspace{-2mm}
\end{table}

\noindent
\textbf{2) How GAP and GSP influences our CFD module?} The mean and standard deviation (std) of each channel are used in our CFD module to estimate the 3D bilateral grid of affine transformation coefficients for exposure correction. To demonstrate their combined effect, we replace the addition of GAP and GSP (denoted as ``GAP + GSP'') in our CFD module with single GAP (denoted as ``GAP'') or singel GSP (denoted as ``GSP'') in our CFD module. As shown in Table \ref{table:2}, with similar inference speed, ``GAP + GSP'' achieves best numerical results, while single GAP performs better than singe GSP. This illustrates that adding the mean and std of each channel in our CFD module is indeed useful. Besides, the mean plays a principal role.
\begin{table}[h]
\vspace{-2mm}  
\setlength{\abovecaptionskip}{3pt}
 \caption{\textbf{Results of only using GAP or GSP in our CFD module.} ``GAP'' (or ``GSP'') means we use only ``GAP'' (or ``GSP'') in our CFD module. ``GAP + GSP'' means we use the method of adding the ``GAP'' and ``GSP'' in our CFD module.}
\vspace{0.3mm}
\label{table:2}
\resizebox{\linewidth}{!}{%
\begin{tabular}{c|cccrrr}
\Xhline{1pt}
\rowcolor[rgb]{ .9,  .9,  .9}
Method & PSNR $\uparrow$ & SSIM $\uparrow$ & LPIPS $\downarrow$ & \# Param. & FLOPs (M) & Speed (ms)\\ \hline
GAP & 20.71 & 0.829 & 0.1688 & 7,594 & 83.73 &4.32 \\
GSP & 20.47 & 0.826 & 0.1670 & 7,594 & 83.17 &4.33 \\
\rowcolor[rgb]{ .95,  .95,  .95}
GAP+GSP & 21.02 & 0.835 & 0.1644 & 7,594 & 83.45 & 4.34  \\ \hline
\end{tabular}%
}
\vspace{-2mm}
\end{table}

\noindent
\textbf{3) How to design and use SFE module in HFD?}. To study this question, we remove SFE modules in HFD or keep only one convolution and ReLU in SFE, denoted as ``w/o SFEs'' and ``w/ Conv-1'', respectively. As shown in Table \ref{table:3}, although removing the SFE module or part of it can reduce parameters and computational costs, the PSNR, SSIM~\cite{ssim} and LPIPS~\cite{zhang2018unreasonable} are not as good as keeping our SFE module.

\begin{table}[h]
\vspace{-2mm}  
\setlength{\abovecaptionskip}{3pt}
 \caption{\textbf{Results of how the SFE modules are present in the HFD module.} ``w/o SFEs'' (``w/ SFEs'') means whether we remove the SFE modules in the HFD. ``w/ Conv-1'' means we replace SFE in HFD module with a simple $1\times1$ convolutional layer and a ReLU layer.}
\vspace{-0.3mm}
\label{table:3}
\resizebox{\linewidth}{!}{%
\begin{tabular}{c|cccrrr}
\Xhline{1pt}
\rowcolor[rgb]{ .9,  .9,  .9}
Method & PSNR $\uparrow$ & SSIM $\uparrow$ & LPIPS $\downarrow$ & \# Param. & FLOPs (M) & Speed (ms)\\ \hline
w/o SFEs & 20.18 & 0.823 & 0.1845 & 2,672 & 60.77 &3.85 \\
w/ Conv-1 & 20.64 & 0.830 & 0.1740 & 4,321 & 72.11 &3.88 \\
\rowcolor[rgb]{ .95,  .95,  .95}
w/ SFEs & 21.02 & 0.835 & 0.1644 & 7,594 & 83.45 & 4.34  \\\hline
\end{tabular}%
}
\vspace{-3mm}
\end{table}

\noindent
\textbf{4) Effect of feature decomposition order in CFD}. Our CFD module decompose the context-aware feature and the residual feature by feature subtraction. Here, we contrast the cases either the context-aware feature or the residual feature used as inputs to the next SFE, respectively. As shown in Table \ref{table:4}, our model performs comparably when the SFE module is fed with the context-aware feature or the residual feature. We conclude that the feature decomposition order in CFD module does not affect the performance of the HFD module.
\begin{table}[h]
\vspace{-2mm}  
\setlength{\abovecaptionskip}{3pt}
 \caption{\textbf{Results of whether the Context-aware feature output by CFDs is input to SFE or Residual feature is input to SFE in HFD module}. ``Context-aware feature'' means we feed the context-aware feature into SFE module and ``Residual feature'' means we feed the residual feature feature into SFE module in our CFD module.}
\vspace{0.5mm}
\label{table:4}
\resizebox{\linewidth}{!}{%
\begin{tabular}{c|cccrrr}
\Xhline{1pt}
\rowcolor[rgb]{ .9,  .9,  .9}
Method & PSNR $\uparrow$ & SSIM $\uparrow$ & LPIPS $\downarrow$ & \# Param. & FLOPs (M) & Speed (ms)\\ \hline
Context-aware feature & 20.81 & 0.827 & 0.1694 & 7,594 & 83.45 &4.35 \\
\rowcolor[rgb]{ .95,  .95,  .95}
Residual feature& 21.02 & 0.835 & 0.1644 & 7,594 & 83.45 & 4.34  \\ \hline
\end{tabular}%
}
\vspace{-5mm}
\end{table}

\subsection{More visual comparisons of our MSLTs with the other comparison methods}
\label{sec:vis1}
Here, we present more visual comparison results with other competing methods on 
the ME dataset~\cite{afifi2021learning} and the SICE~\cite{cai2018learning} dataset here. For the ME dataset, we present two sets of comparison images for each of the five relative exposure values of $\{-1.5,-1,0,+1,+1.5\}$ in Figures~\ref{fig:ev0}-\ref{fig:ev1.5}. For the SICE dataset, we present three sets of comparison images each for under and over exposed inputs in Figures~\ref{fig:sice-under} and \ref{fig:sice-over}. All these results demonstrate that our MSLT networks (MSLT, MSLT+, and MSLT++) achieve comparable or even better visual quality on the exposure corrected images than the competing methods with larger parameter amount and computational costs.

\subsection{Visual comparisons in ablation studies}
\label{sec:vis2}
In this section, we will provide visual comparisons of ablation studies in our paper and this supplementary file. Figures~\ref{fig:LP}-\ref{fig:high} represent the 1st-5th ablation study in our paper and Figures~\ref{fig:share}-\ref{fig:res} represent the 1st-4th ablation study in this supplementary file, respectively. For simplicity, we randomly select one image from the two datasets for comparison in each ablation study.

Specifically, Figure~\ref{fig:LP} shows the visual results of our MSLT with different number of Laplacian pyramid levels on one over-exposure image.  Figures~\ref{fig:CFD} and \ref{fig:CFDnum} show the visual results of our MSLT with different variants of CFD module and different number of CFD modules in HFD. Figure~\ref{fig:suppHFD} shows visual results of our MSLT with different variants of HFD module in the developed Bilateral Grid Network. Figure~\ref{fig:high} shows the visual results of our MSLT and MSLT+ with some high-frequency layers in Laplacian pyramid unprocessed by MSLT/MSLT+. Figure~\ref{fig:share} shows visual results of our MSLT with the parameters of $1\times1$ convolutions shared or not. Figure~\ref{fig:GAP} shows visual results of our MSLT which handles whether or not GAP and GSP are used in CFD module. Figure~\ref{fig:sfe} shows visual results of our MSLT which handles SFE modules differently. Figure~\ref{fig:res} shows visual results of our MSLT with with different inputs to our SFE module. 

As we can see, our MSLT/MSLT+ can better restore the brightness and color of the images than the other methods in all these ablation studies.

\subsection{Societal Impact}
\label{sec:impact}
This work has the potential to be applied to enhance the user experience of taking photos in real-time, and enjoys much positive societal impact.

%


%

\begin{figure*}[t]
    \setlength{\abovecaptionskip}{3pt}
    \centering
    \begin{overpic}[width=0.93\textwidth]{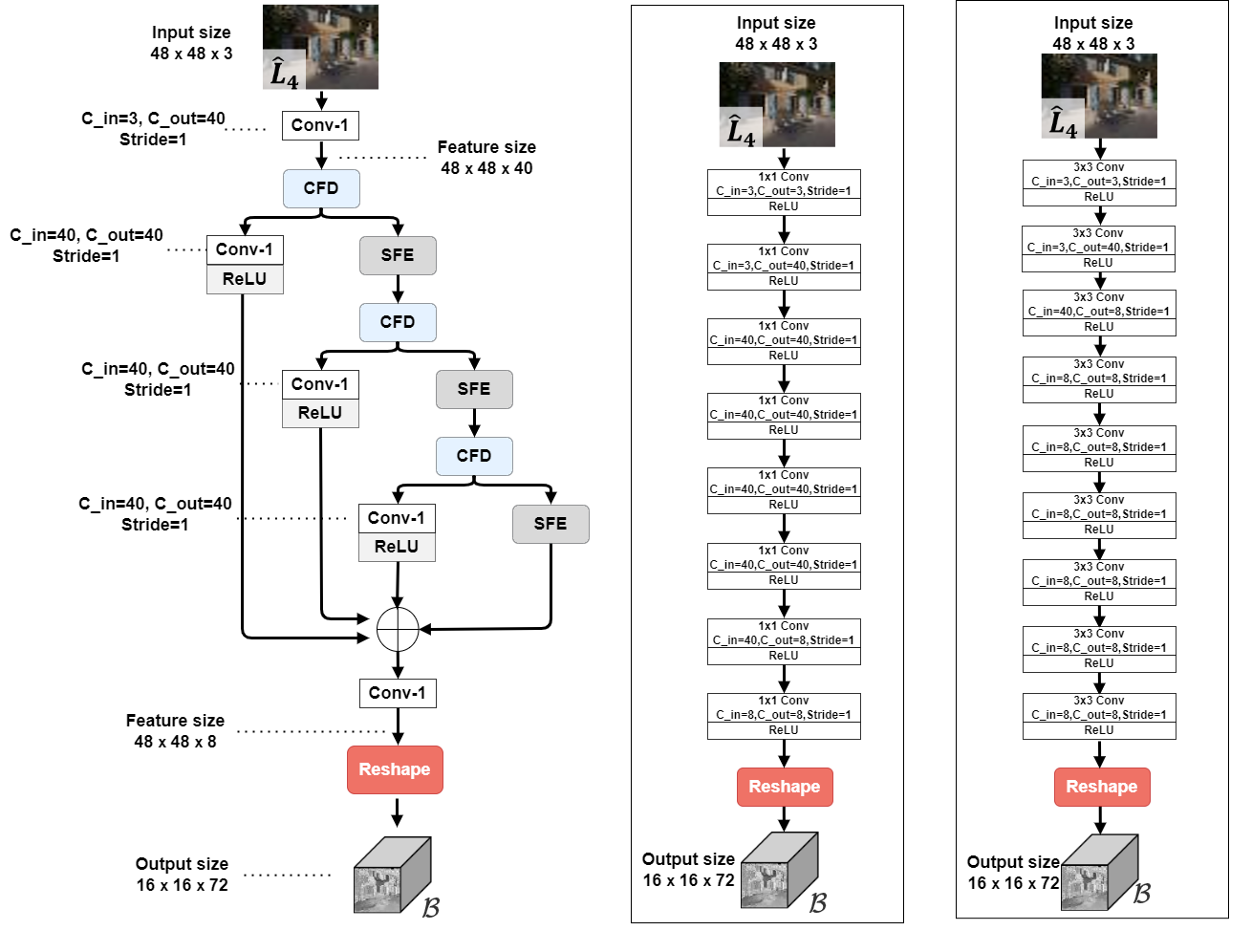}
    \put(20,0){(a) Our HFD module}
    \put(57,0){(b) ``Conv-1''}
    \put(83,0){(c) ``Conv-3''}
    \end{overpic}
    \vspace{0mm}
    \caption{\textbf{The detailed structure of (a) our HFD module, (b) ``Conv-1'' and (c) ``Conv-3'' in Section 4.3 (4) of our paper.} All these three networks take a feature map of $48\times48\times3$ as input and output a 3D bilateral grid of affine coefficients $\mathcal{B}\in\mathbb{R}^{16\times16\times72}$. C{\_}in and C{\_}out denote the number of input and output channels of convolutions, respectively. }
    \label{fig:supp3.4}
    \vspace{-3mm}   
\end{figure*}


\begin{figure*}[h]
    \centering
          \begin{overpic}[width=1\textwidth]{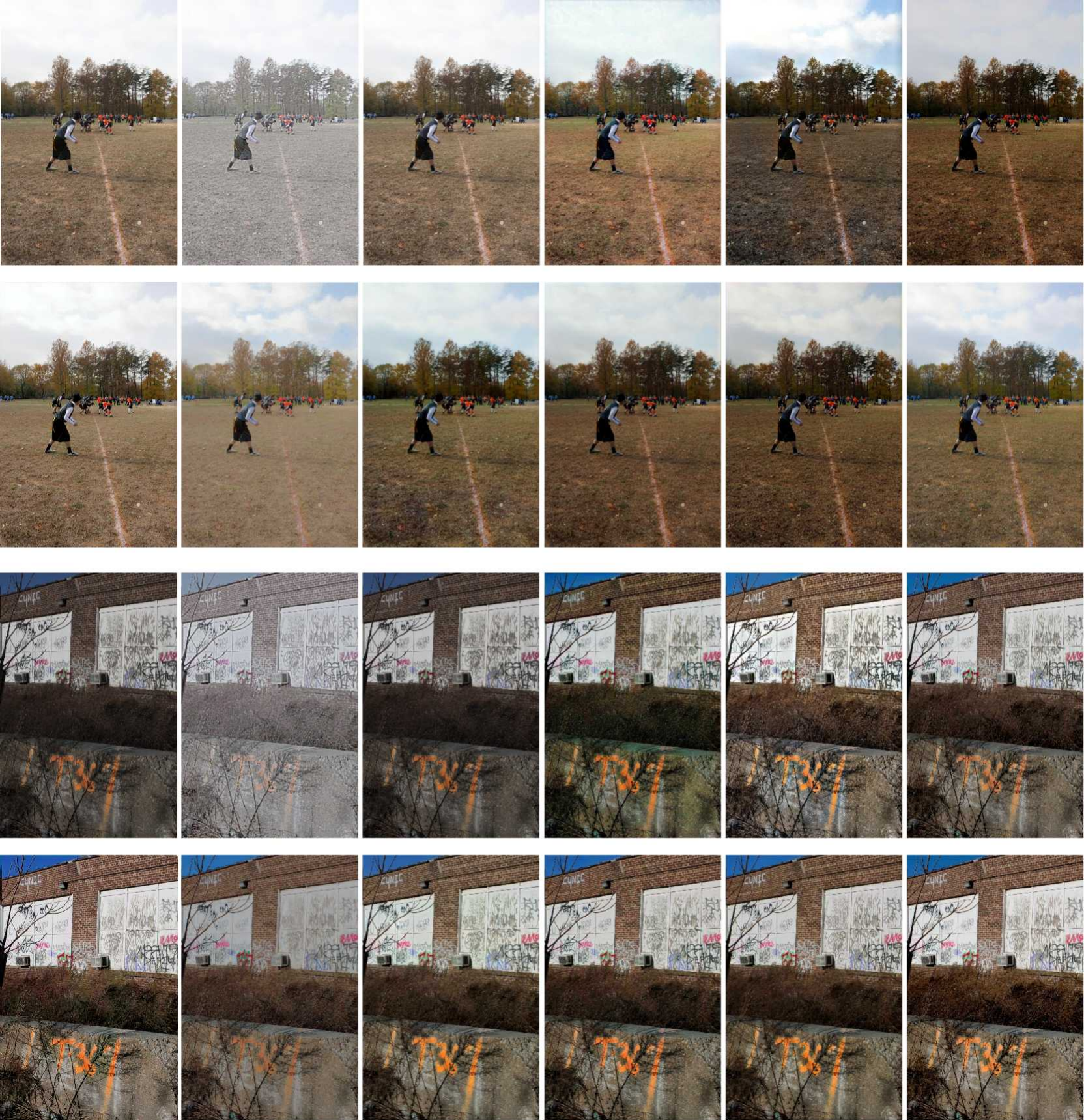}
          \put(6.75,75){\scriptsize Input}
          \put(20.5,75){\scriptsize Zero-DCE~\cite{guo2020zero}}
         \put(38.5,75){\scriptsize SCI~\cite{ma2022toward}}
         \put(54,75){\scriptsize LPTN~\cite{liang2021high}}
         \put(70,75){\scriptsize MSEC~\cite{afifi2021learning}}
        \put(87,75){\scriptsize IAT~\cite{90K}}
   
         \put(6.5,49.85){\scriptsize LCDP \cite{wang2022local}}
        \put(20.5,49.85){\scriptsize Channel-MLP}
        \put(38.5,49.85){\scriptsize MSLT}
        \put(54.5,49.85){\scriptsize MSLT+}
        \put(70,49.85){\scriptsize MSLT++}
        \put(85,49.85){\scriptsize Ground Truth}
         \put(6.75,24){\scriptsize Input}
          \put(20.5,24){\scriptsize Zero-DCE~\cite{guo2020zero}}
         \put(38.5,24){\scriptsize SCI~\cite{ma2022toward}}
         \put(54,24){\scriptsize LPTN~\cite{liang2021high}}
         \put(70,24){\scriptsize MSEC~\cite{afifi2021learning}}
        \put(87,24){\scriptsize IAT~\cite{90K}}
   
         \put(6.5,-1.25){\scriptsize LCDP \cite{wang2022local}}
        \put(20.5,-1.25){\scriptsize Channel-MLP}
        \put(38.5,-1.25){\scriptsize MSLT}
        \put(54.5,-1.25){\scriptsize MSLT+}
        \put(70,-1.25){\scriptsize MSLT++}
        \put(85,-1.25){\scriptsize Ground Truth}
          \end{overpic}
          
    \vspace{2mm}        
    \caption{\textbf{Visual quality comparison of exposure corrected images from ME dataset~\cite{afifi2021learning} for 0 exposure value}.}
    \label{fig:ev0}
    \vspace{-5mm}
\end{figure*}

\begin{figure*}[h]
\vspace{-6mm}
    \centering
          \begin{overpic}[width=1\textwidth]{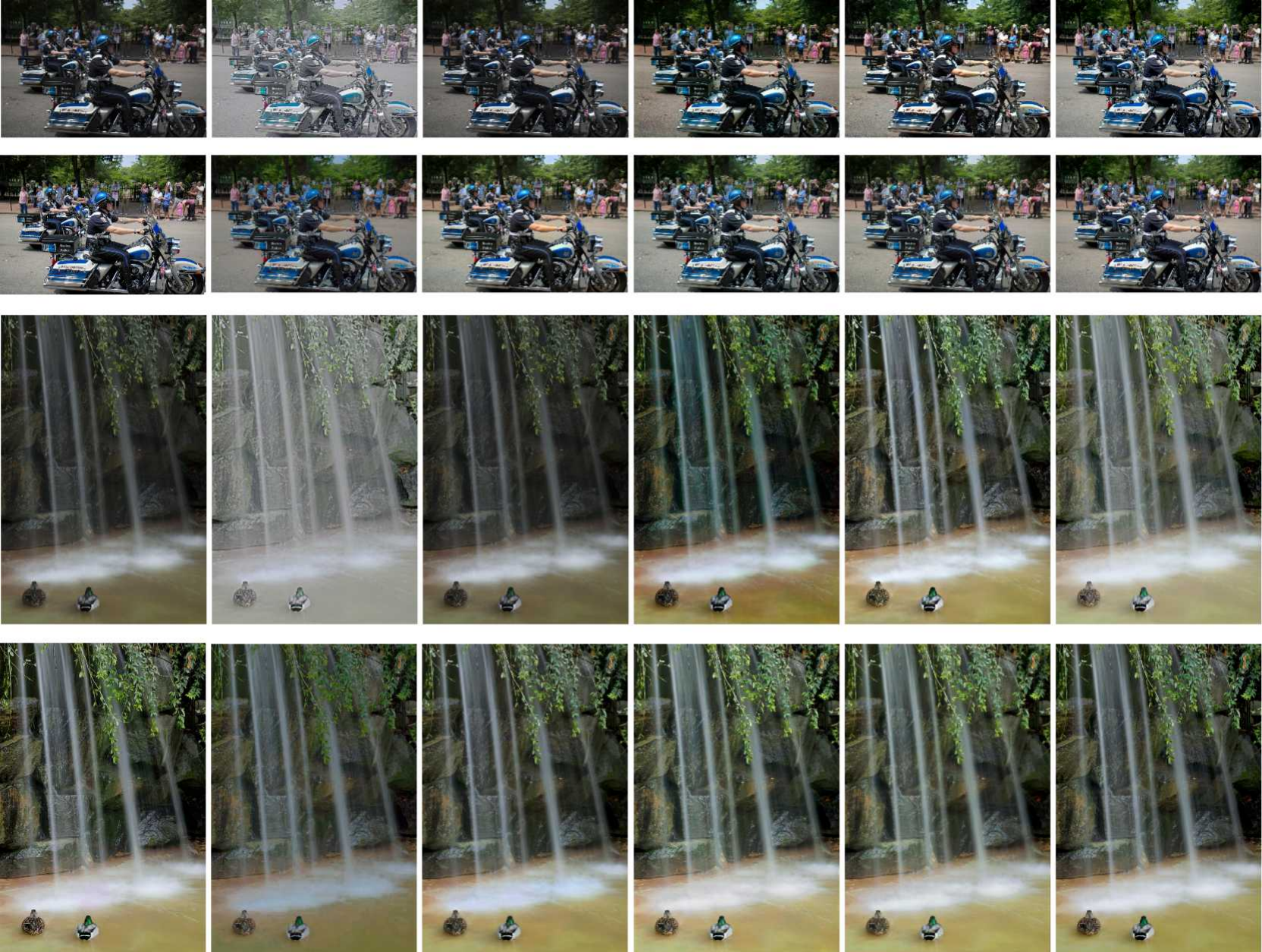}
          \put(6.75,63.25){\scriptsize Input}
          \put(21.25,63.25){\scriptsize Zero-DCE~\cite{guo2020zero}}
         \put(40,63.25){\scriptsize SCI~\cite{ma2022toward}}
         \put(56,63.25){\scriptsize LPTN~\cite{liang2021high}}
         \put(72.5,63.25){\scriptsize MSEC~\cite{afifi2021learning}}
        \put(90,63.25){\scriptsize IAT~\cite{90K}}
   
         \put(6.5,50.85){\scriptsize LCDP \cite{wang2022local}}
        \put(21,50.85){\scriptsize Channel-MLP}
        \put(40,50.85){\scriptsize MSLT}
        \put(56.5,50.85){\scriptsize MSLT+}
        \put(73,50.85){\scriptsize MSLT++}
        \put(88,50.85){\scriptsize Ground Truth}
         \put(6.75,25){\scriptsize Input}
          \put(21.25,25){\scriptsize Zero-DCE~\cite{guo2020zero}}
         \put(40,25){\scriptsize SCI~\cite{ma2022toward}}
         \put(56,25){\scriptsize LPTN~\cite{liang2021high}}
         \put(72.5,25){\scriptsize MSEC~\cite{afifi2021learning}}
        \put(90,25){\scriptsize IAT~\cite{90K}}
   
         \put(6.5,-1.25){\scriptsize LCDP \cite{wang2022local}}
        \put(21,-1.25){\scriptsize Channel-MLP}
        \put(40,-1.25){\scriptsize MSLT}
        \put(56.5,-1.25){\scriptsize MSLT+}
        \put(73,-1.25){\scriptsize MSLT++}
        \put(88,-1.25){\scriptsize Ground Truth}
          \end{overpic}
    \vspace{-5mm}        
    \caption{\textbf{Visual quality comparison of exposure corrected images from ME dataset~\cite{afifi2021learning} for -1 exposure value}.}
    \label{fig:ev-1}
\end{figure*}

%

\begin{figure*}[h]
\vspace{-4mm}
    \centering
          \begin{overpic}[width=1\textwidth]{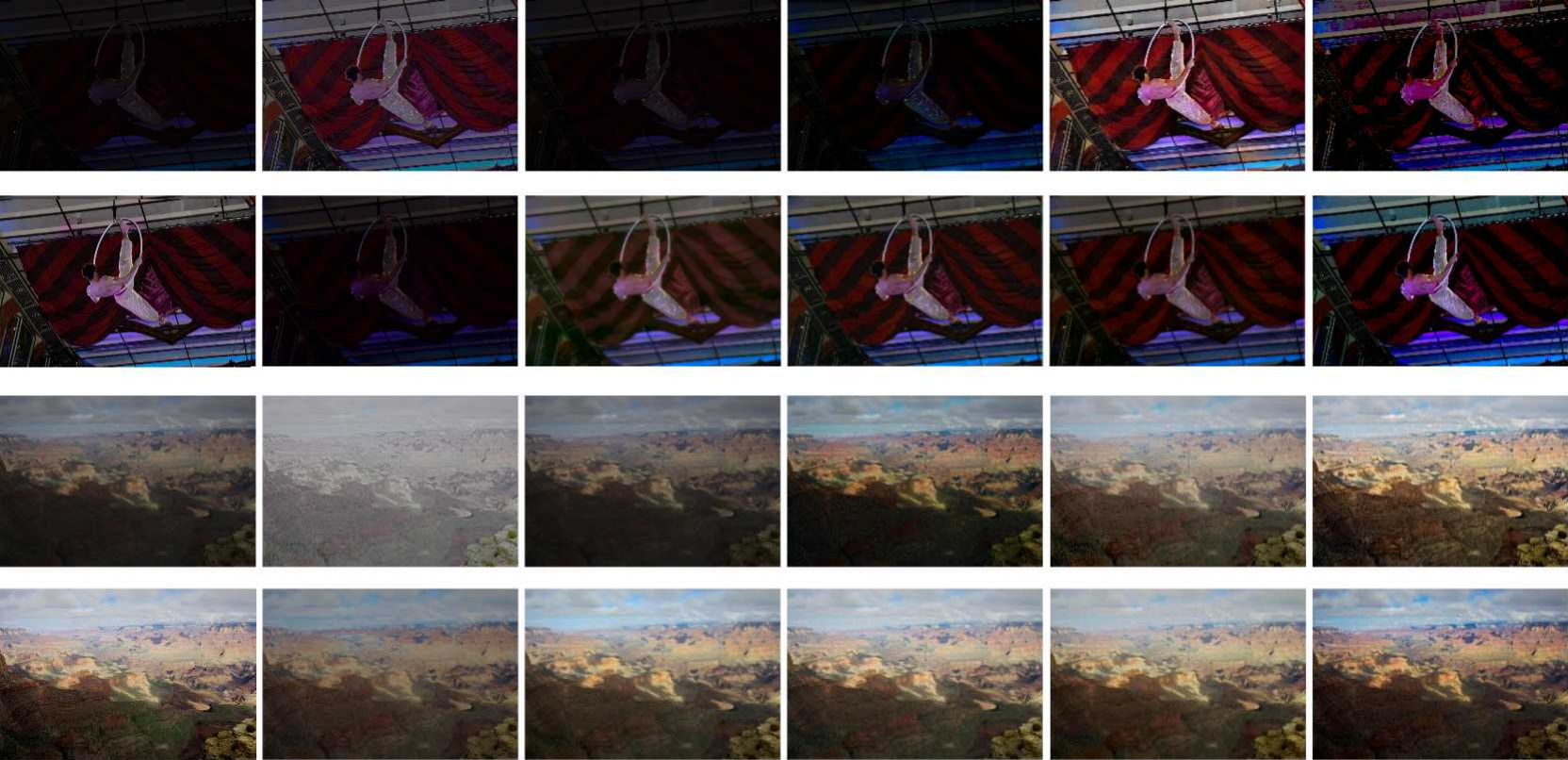}
          \put(6.75,36.25){\scriptsize Input}
          \put(21.25,36.25){\scriptsize Zero-DCE~\cite{guo2020zero}}
         \put(40,36.25){\scriptsize SCI~\cite{ma2022toward}}
         \put(56,36.25){\scriptsize LPTN~\cite{liang2021high}}
         \put(72.5,36.25){\scriptsize MSEC~\cite{afifi2021learning}}
        \put(90,36.25){\scriptsize IAT~\cite{90K}}
   
         \put(6.5,23.85){\scriptsize LCDP \cite{wang2022local}}
        \put(21,23.85){\scriptsize Channel-MLP}
        \put(40,23.85){\scriptsize MSLT}
        \put(56.5,23.85){\scriptsize MSLT+}
        \put(73,23.85){\scriptsize MSLT++}
        \put(88,23.85){\scriptsize Ground Truth}
         \put(6.75,11.25){\scriptsize Input}
          \put(21.25,11.25){\scriptsize Zero-DCE~\cite{guo2020zero}}
         \put(40,11.25){\scriptsize SCI~\cite{ma2022toward}}
         \put(56,11.25){\scriptsize LPTN~\cite{liang2021high}}
         \put(72.5,11.25){\scriptsize MSEC~\cite{afifi2021learning}}
        \put(90,11.25){\scriptsize IAT~\cite{90K}}
   
         \put(6.5,-1.25){\scriptsize LCDP \cite{wang2022local}}
        \put(21,-1.25){\scriptsize Channel-MLP}
        \put(40,-1.25){\scriptsize MSLT}
        \put(56.5,-1.25){\scriptsize MSLT+}
        \put(73,-1.25){\scriptsize MSLT++}
        \put(88,-1.25){\scriptsize Ground Truth}
          \end{overpic}
    \vspace{-4mm}        
    \caption{\textbf{Visual quality comparison of exposure corrected images from ME dataset~\cite{afifi2021learning} for -1.5 exposure value}.}
    \label{fig:ev-1.5}
    \vspace{-1mm}
\end{figure*}


\begin{figure*}[h]
\vspace{-5mm}
    \centering
          \begin{overpic}[width=1\textwidth]{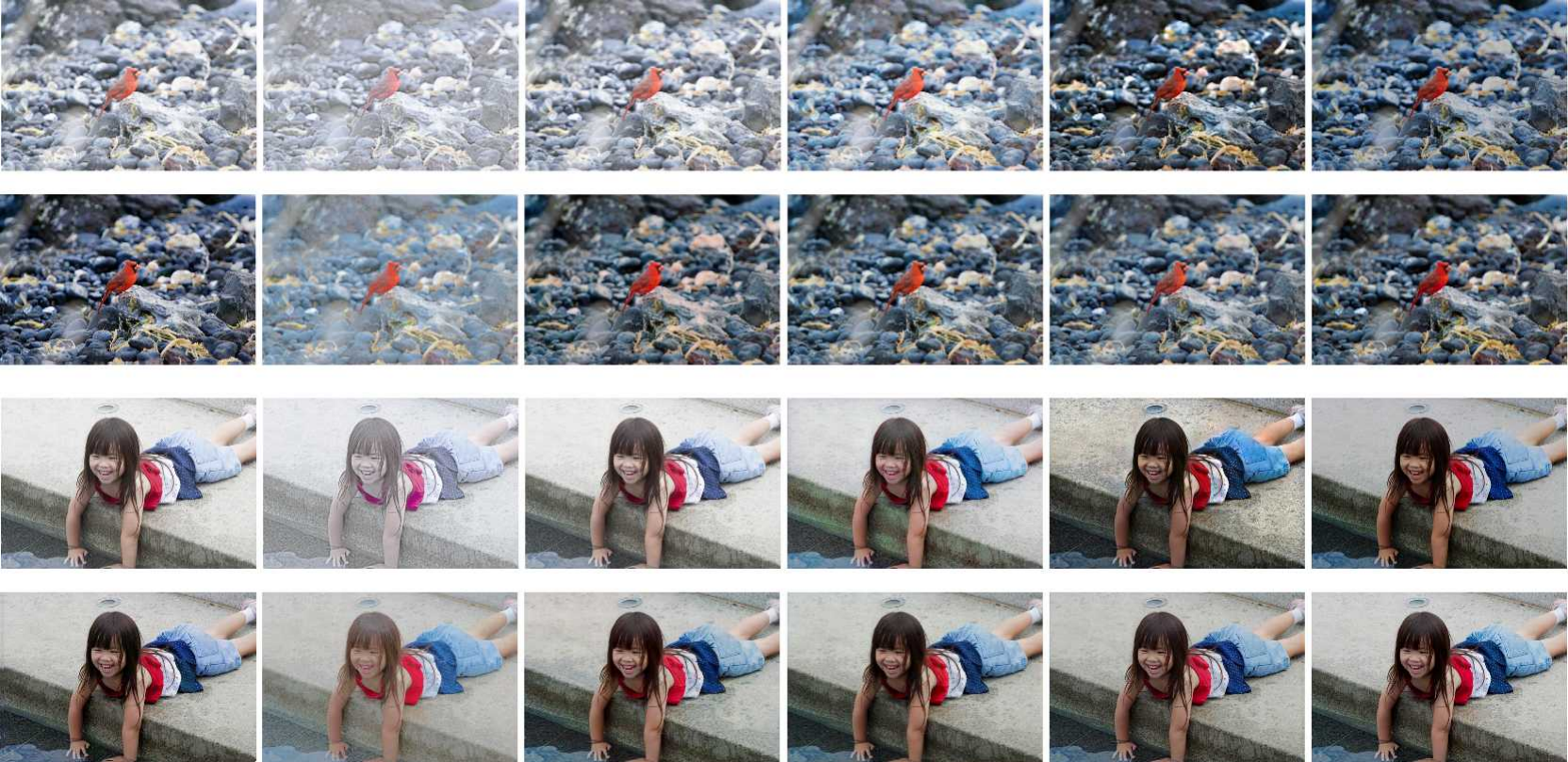}
          \put(6.75,36.25){\scriptsize Input}
          \put(21.25,36.25){\scriptsize Zero-DCE~\cite{guo2020zero}}
         \put(40,36.25){\scriptsize SCI~\cite{ma2022toward}}
         \put(56,36.25){\scriptsize LPTN~\cite{liang2021high}}
         \put(72.5,36.25){\scriptsize MSEC~\cite{afifi2021learning}}
        \put(90,36.25){\scriptsize IAT~\cite{90K}}
   
         \put(6.5,23.85){\scriptsize LCDP \cite{wang2022local}}
        \put(21,23.85){\scriptsize Channel-MLP}
        \put(40,23.85){\scriptsize MSLT}
        \put(56.5,23.85){\scriptsize MSLT+}
        \put(73,23.85){\scriptsize MSLT++}
        \put(88,23.85){\scriptsize Ground Truth}
         \put(6.75,11.25){\scriptsize Input}
          \put(21.25,11.25){\scriptsize Zero-DCE~\cite{guo2020zero}}
         \put(40,11.25){\scriptsize SCI~\cite{ma2022toward}}
         \put(56,11.25){\scriptsize LPTN~\cite{liang2021high}}
         \put(72.5,11.25){\scriptsize MSEC~\cite{afifi2021learning}}
        \put(90,11.25){\scriptsize IAT~\cite{90K}}
   
         \put(6.5,-1.25){\scriptsize LCDP \cite{wang2022local}}
        \put(21,-1.25){\scriptsize Channel-MLP}
        \put(40,-1.25){\scriptsize MSLT}
        \put(56.5,-1.25){\scriptsize MSLT+}
        \put(73,-1.25){\scriptsize MSLT++}
        \put(88,-1.25){\scriptsize Ground Truth}
          \end{overpic}
    \vspace{-3mm}        
    \caption{\textbf{Visual quality comparison of exposure corrected images from ME dataset~\cite{afifi2021learning} for +1 exposure value}.}
    \label{fig:ev1}
\end{figure*}


\begin{figure*}[h]
    \centering
          \begin{overpic}[width=1\textwidth]{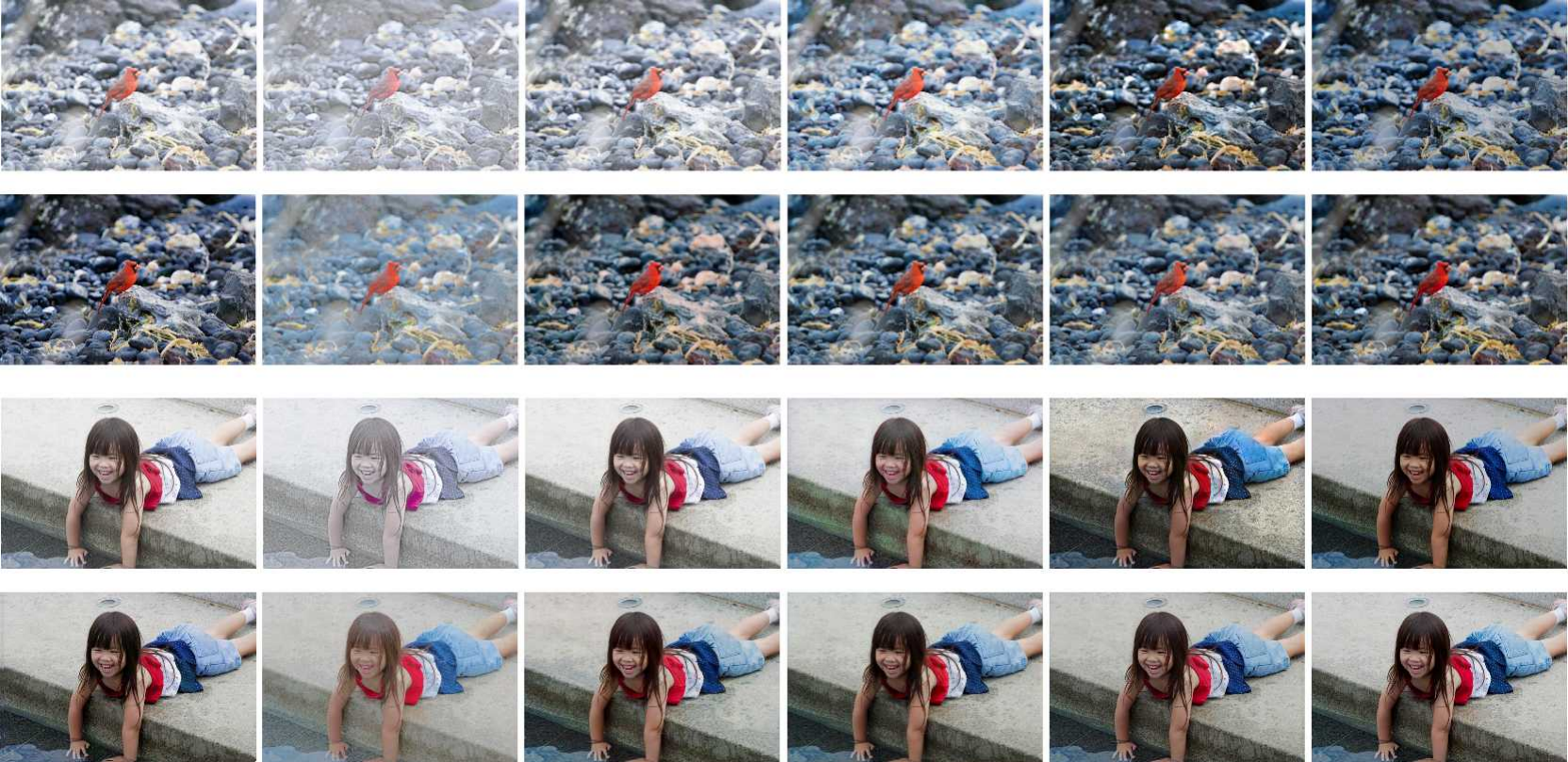}
          \put(6.75,36.75){\scriptsize Input}
          \put(21.25,36.75){\scriptsize Zero-DCE~\cite{guo2020zero}}
         \put(40,36.75){\scriptsize SCI~\cite{ma2022toward}}
         \put(56,36.75){\scriptsize LPTN~\cite{liang2021high}}
         \put(72.5,36.75){\scriptsize MSEC~\cite{afifi2021learning}}
        \put(90,36.75){\scriptsize IAT~\cite{90K}}
   
         \put(6.5,23.85){\scriptsize LCDP \cite{wang2022local}}
        \put(21,23.85){\scriptsize Channel-MLP}
        \put(40,23.85){\scriptsize MSLT}
        \put(56.5,23.85){\scriptsize MSLT+}
        \put(73,23.85){\scriptsize MSLT++}
        \put(88,23.85){\scriptsize Ground Truth}
         \put(6.75,11.25){\scriptsize Input}
          \put(21.25,11.25){\scriptsize Zero-DCE~\cite{guo2020zero}}
         \put(40,11.25){\scriptsize SCI~\cite{ma2022toward}}
         \put(56,11.25){\scriptsize LPTN~\cite{liang2021high}}
         \put(72.5,11.25){\scriptsize MSEC~\cite{afifi2021learning}}
        \put(90,11.25){\scriptsize IAT~\cite{90K}}
   
         \put(6.5,-1.25){\scriptsize LCDP \cite{wang2022local}}
        \put(21,-1.25){\scriptsize Channel-MLP}
        \put(40,-1.25){\scriptsize MSLT}
        \put(56.5,-1.25){\scriptsize MSLT+}
        \put(73,-1.25){\scriptsize MSLT++}
        \put(88,-1.25){\scriptsize Ground Truth}
          \end{overpic}
    \vspace{-3mm}        
    \caption{\textbf{Visual quality comparison of exposure corrected images from ME dataset~\cite{afifi2021learning} for +1.5 exposure value}.}
    \label{fig:ev1.5}
    \vspace{-5mm}
\end{figure*}


\begin{figure*}[h]
    \centering
     \vspace{+2mm}
          \begin{overpic}[width=1\textwidth]{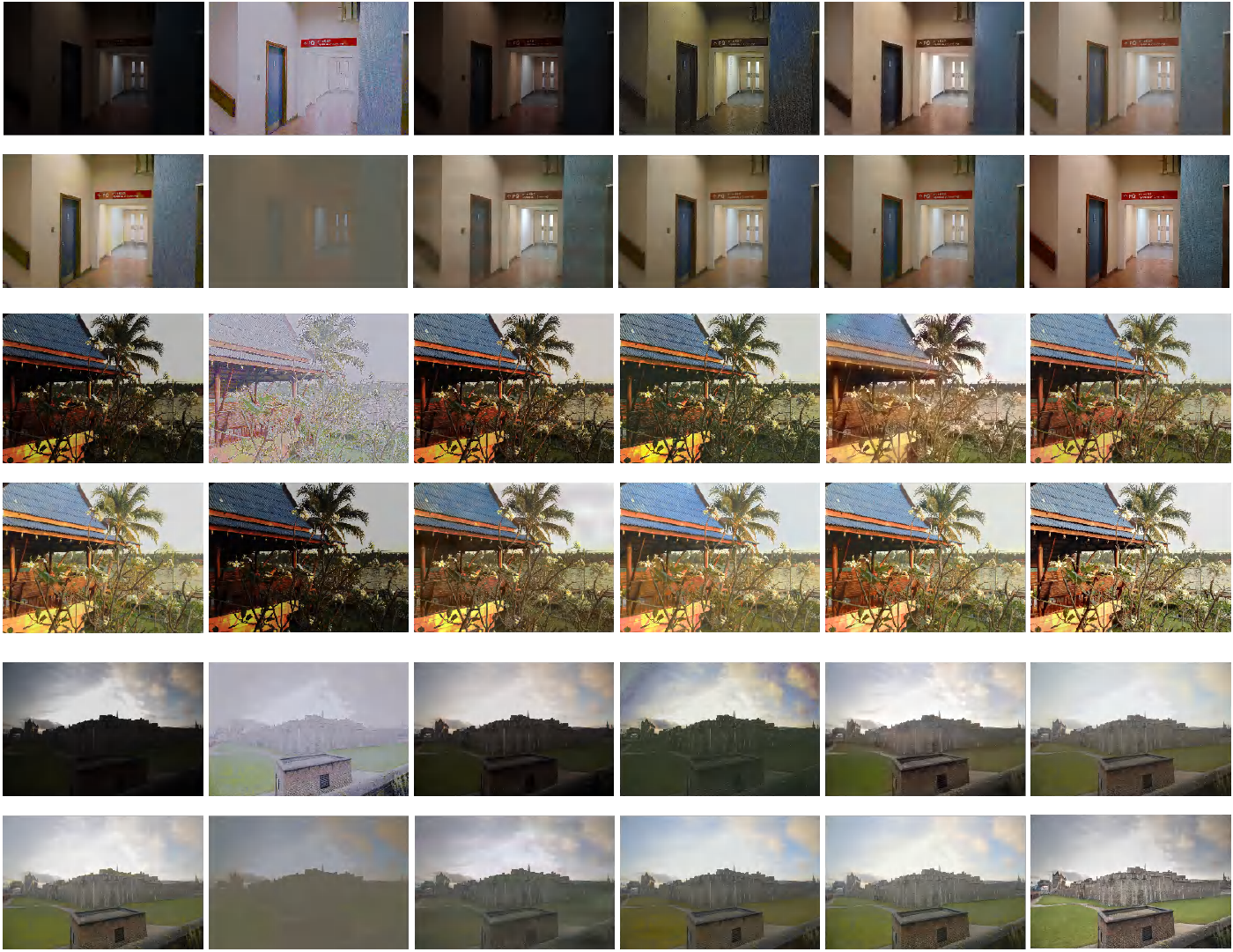}
          \put(6.75,65.25){\scriptsize Input}
          \put(21,65.25){\scriptsize Zero-DCE~\cite{guo2020zero}}
         \put(39.25,65.25){\scriptsize SCI~\cite{ma2022toward}}
         \put(55.5,65.25){\scriptsize LPTN~\cite{liang2021high}}
         \put(72.5,65.25){\scriptsize MSEC~\cite{afifi2021learning}}
        \put(89.5,65.25){\scriptsize IAT~\cite{90K}}
   
         \put(5.5,52.55){\scriptsize LCDP \cite{wang2022local}}
        \put(21,52.55){\scriptsize Channel-MLP}
        \put(40,52.55){\scriptsize MSLT}
        \put(56.5,52.55){\scriptsize MSLT+}
        \put(73,52.55){\scriptsize MSLT++}
        \put(88,52.55){\scriptsize Ground Truth}
         \put(6.75,38.4){\scriptsize Input}
          \put(21,38.35){\scriptsize Zero-DCE~\cite{guo2020zero}}
         \put(39.25,38.35){\scriptsize SCI~\cite{ma2022toward}}
         \put(55.5,38.35){\scriptsize LPTN~\cite{liang2021high}}
         \put(72.5,38.35){\scriptsize MSEC~\cite{afifi2021learning}}
        \put(89.5,38.35){\scriptsize IAT~\cite{90K}}
   
         \put(5.5,24.5){\scriptsize LCDP \cite{wang2022local}}
        \put(21,24.5){\scriptsize Channel-MLP}
        \put(40,24.5){\scriptsize MSLT}
        \put(56.5,24.5){\scriptsize MSLT+}
        \put(73,24.5){\scriptsize MSLT++}
        \put(88,24.5){\scriptsize Ground Truth}
        \put(6.75,11.4){\scriptsize Input}
         \put(21,11.3){\scriptsize Zero-DCE~\cite{guo2020zero}}
         \put(39.25,11.3){\scriptsize SCI~\cite{ma2022toward}}
         \put(55.5,11.3){\scriptsize LPTN~\cite{liang2021high}}
         \put(72.5,11.3){\scriptsize MSEC~\cite{afifi2021learning}}
        \put(89.5,11.3){\scriptsize IAT~\cite{90K}}
   
         \put(5.5,-1.5){\scriptsize LCDP \cite{wang2022local}}
        \put(21,-1.5){\scriptsize Channel-MLP}
        \put(40,-1.5){\scriptsize MSLT}
        \put(56.5,-1.5){\scriptsize MSLT+}
        \put(73,-1.5){\scriptsize MSLT++}
        \put(88,-1.5){\scriptsize Ground Truth}
          \end{overpic}
          
    \vspace{1mm}        
    \caption{\textbf{Visual quality comparison of under  exposure corrected images from SICE dataset~\cite{cai2018learning}}.}
    \label{fig:sice-under}
\end{figure*}

\begin{figure*}[h]
    \centering
     \vspace{+2mm}
          \begin{overpic}[width=1\textwidth]{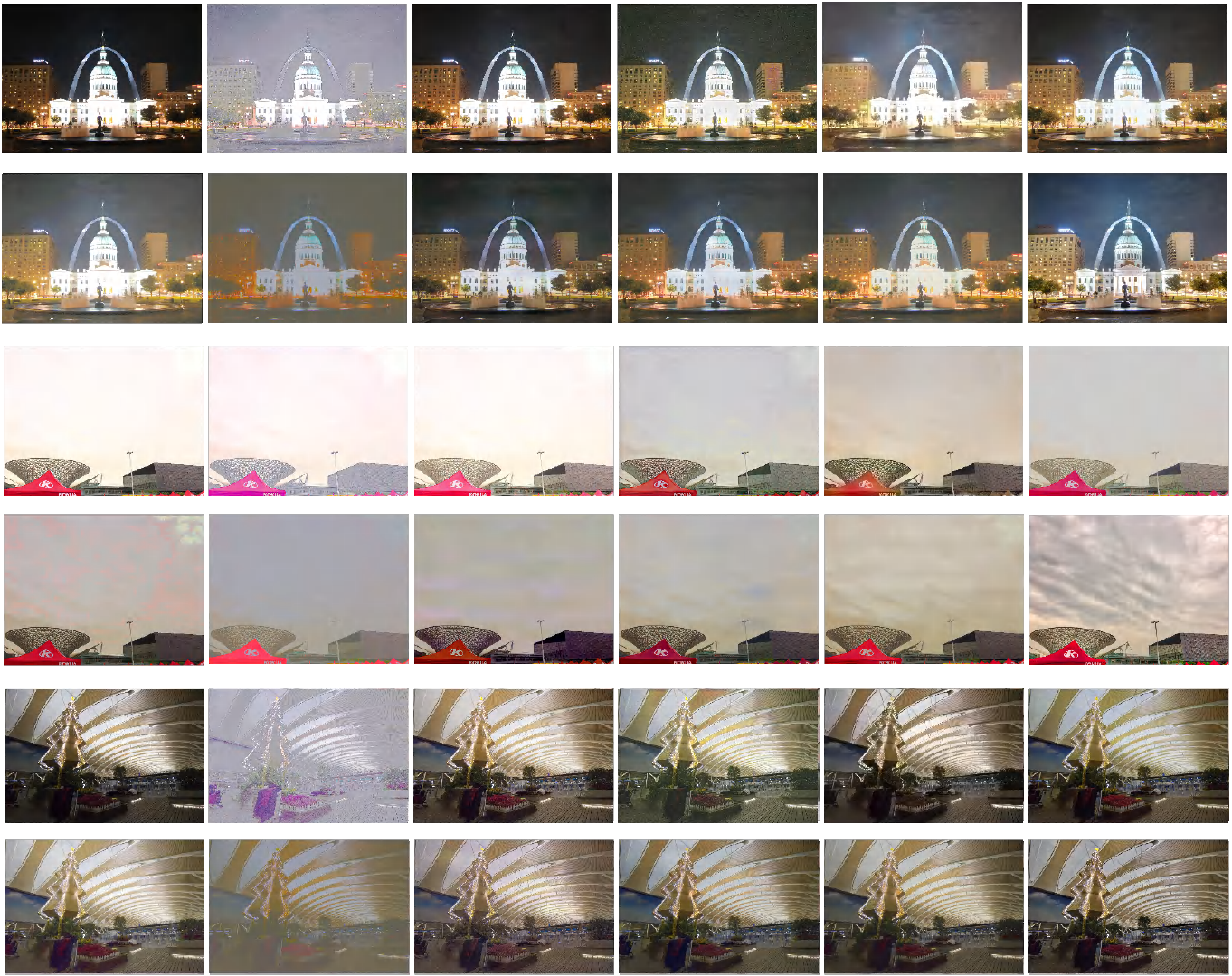}
          \put(6.75,65.75){\scriptsize Input}
          \put(21,65.75){\scriptsize Zero-DCE~\cite{guo2020zero}}
         \put(39.25,65.75){\scriptsize SCI~\cite{ma2022toward}}
         \put(55.5,65.75){\scriptsize LPTN~\cite{liang2021high}}
         \put(72.5,65.75){\scriptsize MSEC~\cite{afifi2021learning}}
        \put(89.5,65.75){\scriptsize IAT~\cite{90K}}
   
         \put(5.5,51.9){\scriptsize LCDP \cite{wang2022local}}
        \put(21,51.9){\scriptsize Channel-MLP}
        \put(40,51.9){\scriptsize MSLT}
        \put(56.5,51.9){\scriptsize MSLT+}
        \put(73,51.9){\scriptsize MSLT++}
        \put(88,51.9){\scriptsize Ground Truth}
         \put(6.75,38){\scriptsize Input}
          \put(21,38){\scriptsize Zero-DCE~\cite{guo2020zero}}
         \put(39.25,38){\scriptsize SCI~\cite{ma2022toward}}
         \put(55.5,38){\scriptsize LPTN~\cite{liang2021high}}
         \put(72.5,38){\scriptsize MSEC~\cite{afifi2021learning}}
        \put(89.5,38){\scriptsize IAT~\cite{90K}}
   
         \put(5.5,24.2){\scriptsize LCDP \cite{wang2022local}}
        \put(21,24.2){\scriptsize Channel-MLP}
        \put(40,24.2){\scriptsize MSLT}
        \put(56.5,24.2){\scriptsize MSLT+}
        \put(73,24.2){\scriptsize MSLT++}
        \put(88,24.2){\scriptsize Ground Truth}
        \put(6.75,11.3){\scriptsize Input}
         \put(21,11.3){\scriptsize Zero-DCE~\cite{guo2020zero}}
         \put(39.25,11.3){\scriptsize SCI~\cite{ma2022toward}}
         \put(55.5,11.3){\scriptsize LPTN~\cite{liang2021high}}
         \put(72.5,11.3){\scriptsize MSEC~\cite{afifi2021learning}}
        \put(89.5,11.3){\scriptsize IAT~\cite{90K}}
   
         \put(5.5,-1.25){\scriptsize LCDP \cite{wang2022local}}
        \put(21,-1.25){\scriptsize Channel-MLP}
        \put(40,-1.25){\scriptsize MSLT}
        \put(56.5,-1.25){\scriptsize MSLT+}
        \put(73,-1.25){\scriptsize MSLT++}
        \put(88,-1.25){\scriptsize Ground Truth}
          \end{overpic}
          
    \vspace{1mm}        
    \caption{\textbf{Visual quality comparison of over  exposure corrected images from SICE dataset~\cite{cai2018learning}}.}
    \label{fig:sice-over}
\end{figure*}


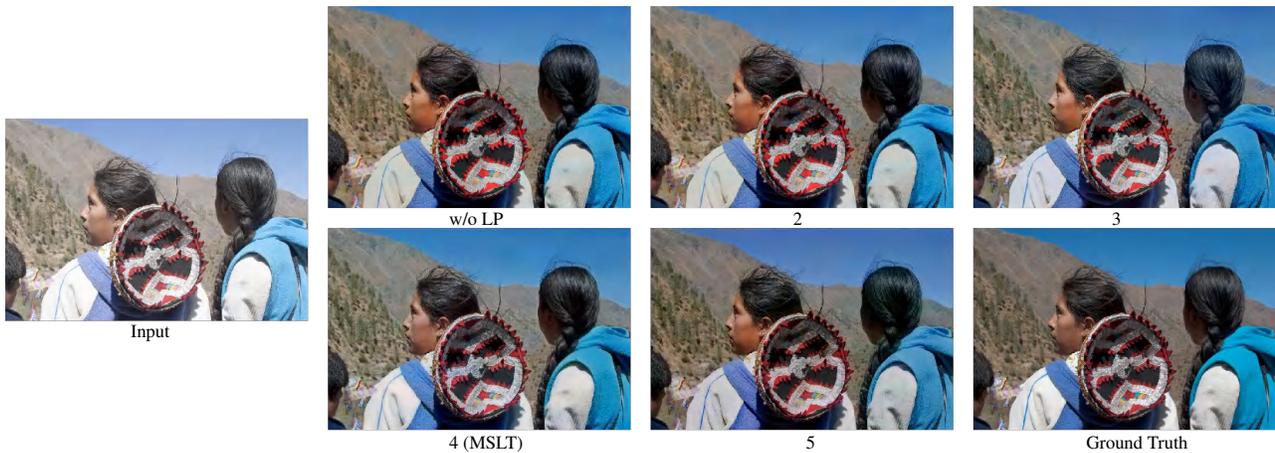
\begin{figure*}[h]
    \setlength{\abovecaptionskip}{3pt}
    \centering
    \begin{overpic}[width=0.97\textwidth]{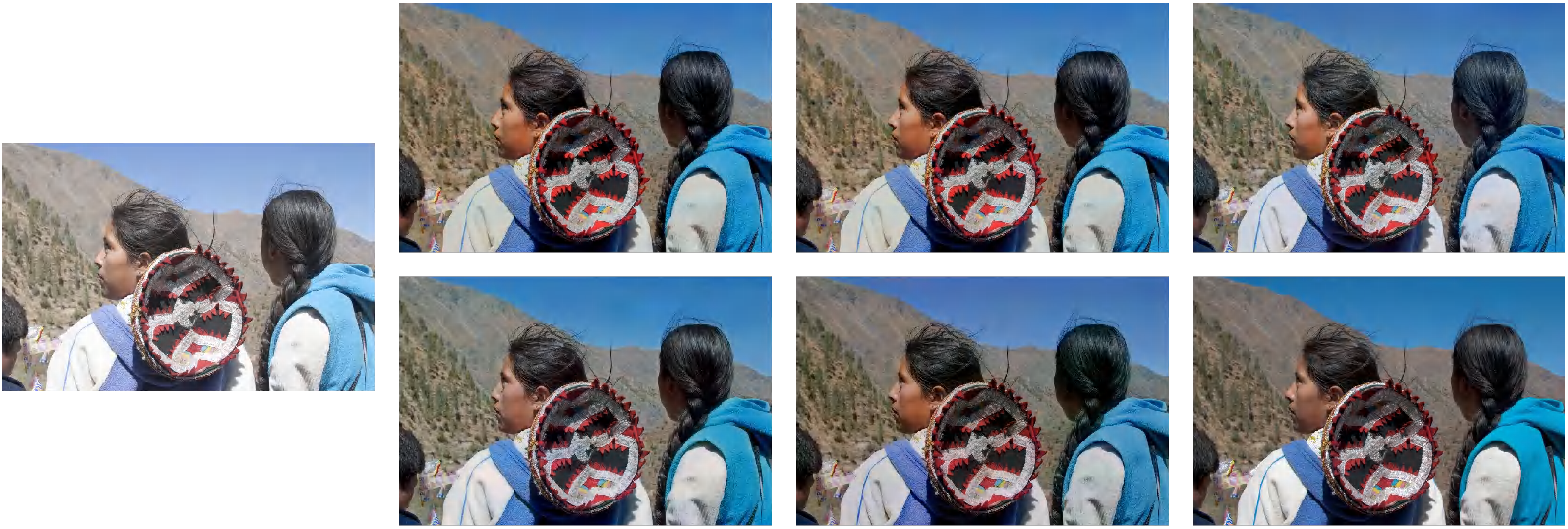}
    
      \put(10,7.5){\scriptsize Input}
      \put(35,16.35){\scriptsize w/o\ LP}
      \put(62,16.35){\scriptsize 2}
      \put(87,16.35){\scriptsize 3}
      \put(35,-1.25){\scriptsize 4 (MSLT)}
      \put(63,-1.25){\scriptsize 5}
      \put(85,-1.25){\scriptsize Ground Truth}
    \end{overpic}
    \vspace{1.5mm}
    \caption{\textbf{Visual quality comparison of exposure corrected images processed by our MSLT with different number of Laplacian pyramid levels}. ``w/o\ LP'' means we do not use Laplacian pyramid.}
    \label{fig:LP}
    \vspace{-2mm}   
\end{figure*}


\begin{figure*}[h]
    \setlength{\abovecaptionskip}{3pt}
    \centering
    \begin{overpic}[width=1\textwidth]{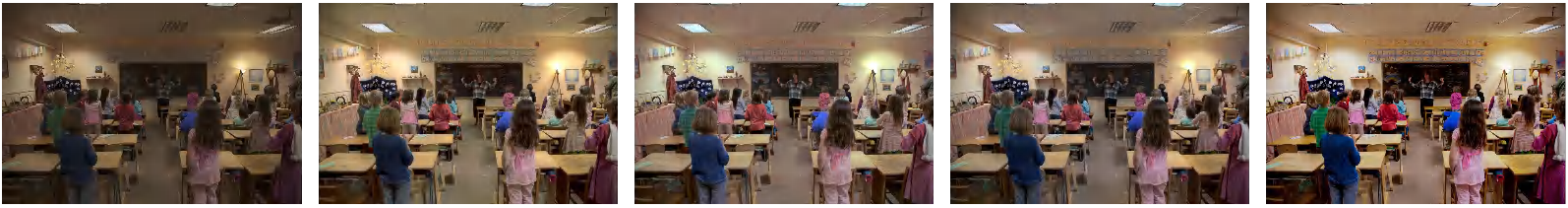}
     \put(8.5,-1.5){\footnotesize Input}
      \put(29.5,-1.5){\footnotesize IN}
      \put(49,-1.5){\footnotesize CA}
      \put(66,-1.5){\footnotesize CFD (MSLT)}
      \put(86,-1.5){\footnotesize Ground Truth}
    \end{overpic}
    \vspace{-2mm}
    \caption{\textbf{Visual quality comparison of exposure corrected images processed by our MSLT with different variants of CFD module in our HFD module}. ``CFD": Context-aware Feature Decomposition. ``IN": Instance Normalization~\cite{ulyanov2016instance} with feature decomposition. ``CA": Channel Attention~\cite{hu2018squeeze} with feature decomposition.}
    \label{fig:CFD}
    \vspace{-2mm}   
\end{figure*}
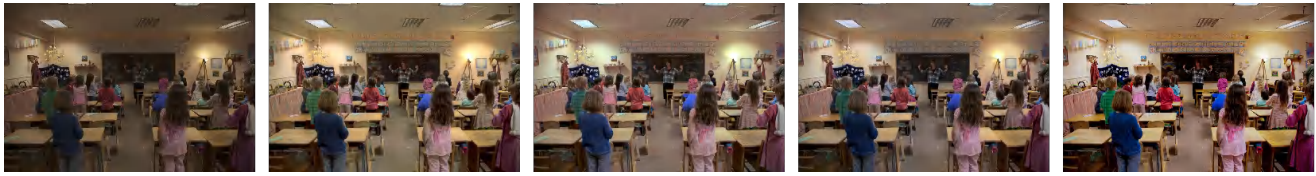


\begin{figure*}[h]
    \setlength{\abovecaptionskip}{3pt}
    \centering
    \begin{overpic}[width=0.97\textwidth]{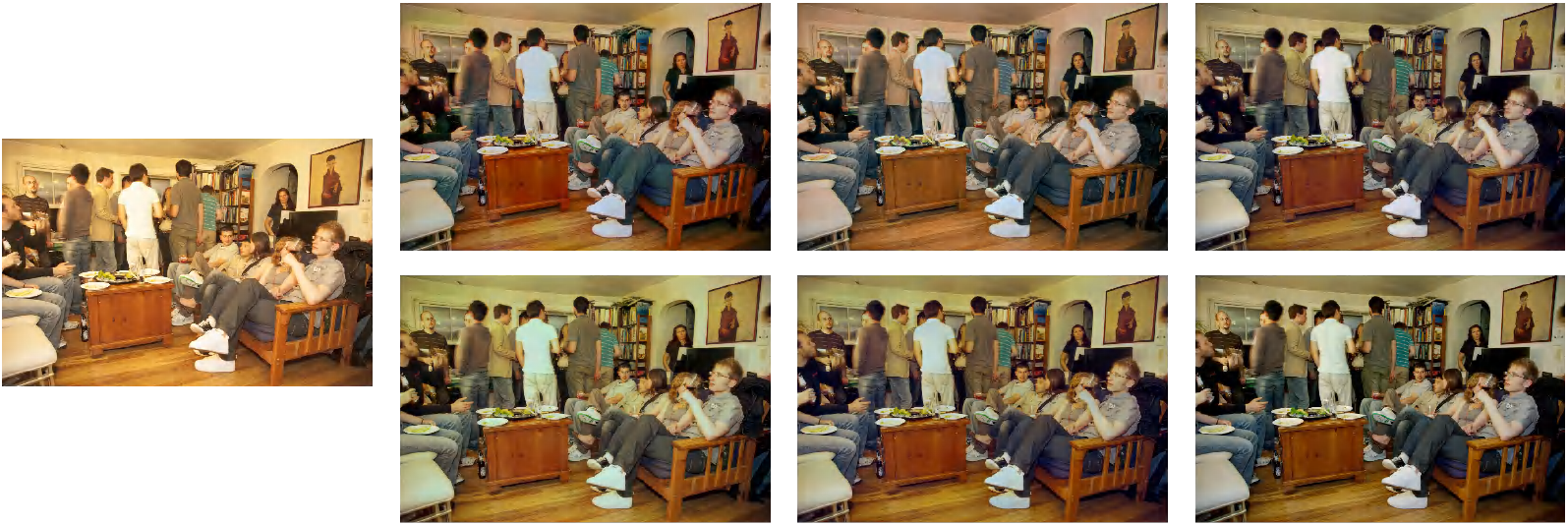}
      \put(10,8){\scriptsize Input}
      \put(37,16.35){\scriptsize 1}
      \put(62,16.35){\scriptsize 2}
      \put(87,16.35){\scriptsize 3 (MSLT)}
      \put(37,-1.25){\scriptsize 4}
      \put(63,-1.25){\scriptsize 5}
      \put(85,-1.25){\scriptsize Ground Truth}
    \end{overpic}
    \vspace{1mm}
    \caption{\textbf{Visual quality comparison of exposure corrected images processed by our MSLT  with different number of CFD modules} in the proposed HFD module.}
    \label{fig:CFDnum}
    \vspace{-2mm}   
\end{figure*}
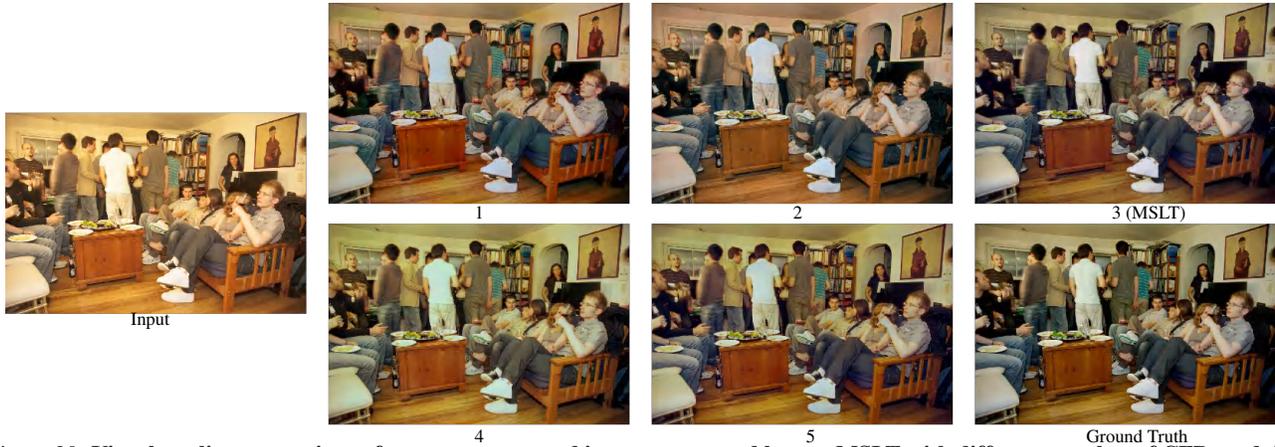


\begin{figure*}[h]
    \setlength{\abovecaptionskip}{3pt}
    \centering
    \begin{overpic}[width=1\textwidth]{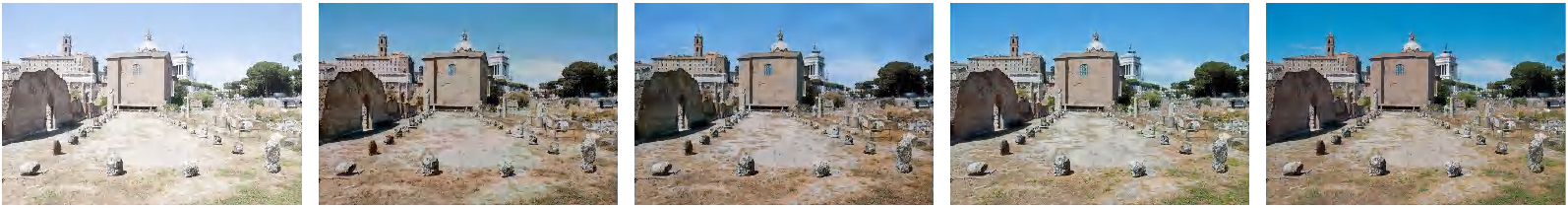}
     \put(8,-1.5){\footnotesize Input}
      \put(28.5,-1.5){\footnotesize Conv-1}
      \put(48,-1.5){\footnotesize Conv-3}
      \put(66,-1.5){\footnotesize HFD (MSLT)}
      \put(87,-1.5){\footnotesize Ground Truth}
    \end{overpic}
    \vspace{-2mm}
    \caption{\textbf{Visual quality comparison of exposure corrected images processed by our MSLT with different variants of HFD module} in the developed Bilateral Grid Network. ``Conv-1" (or ``Conv-3"): the network consisting of multiple $1\times1$ (or $3\times3$) convolutional layers and ReLU activation function(see \ref{fig:supp3.4}). ``HFD": our Hierarchical Feature Decomposition module.}
    \label{fig:suppHFD}
    \vspace{-2mm}   
\end{figure*}
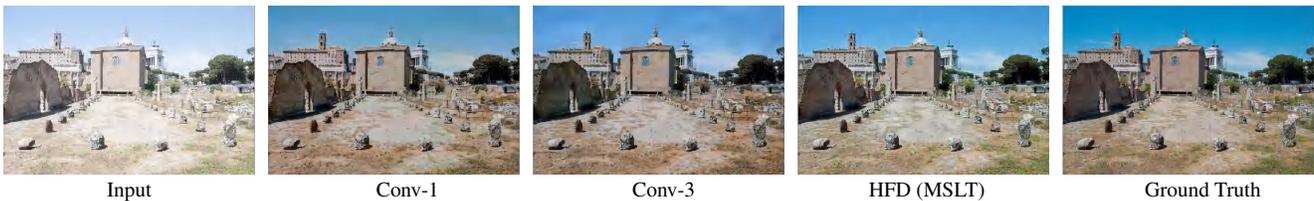


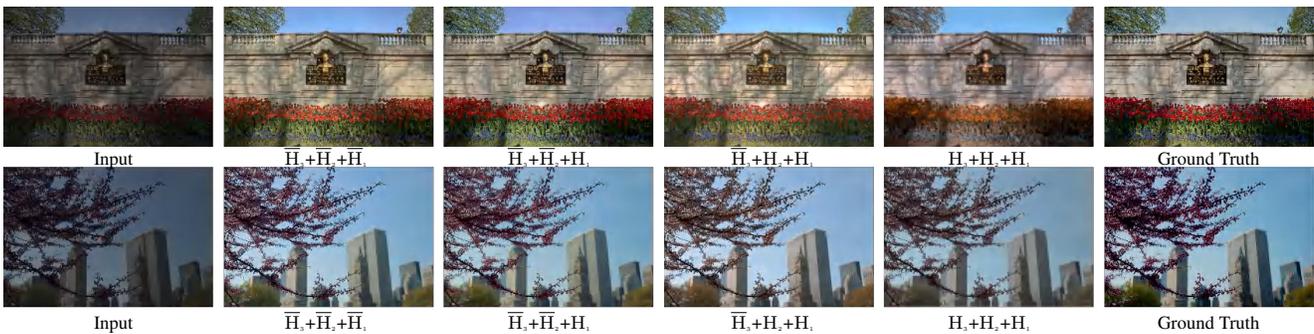
\begin{figure*}[h]
    \setlength{\abovecaptionskip}{3pt}
    \centering
    \begin{overpic}[width=1\textwidth]{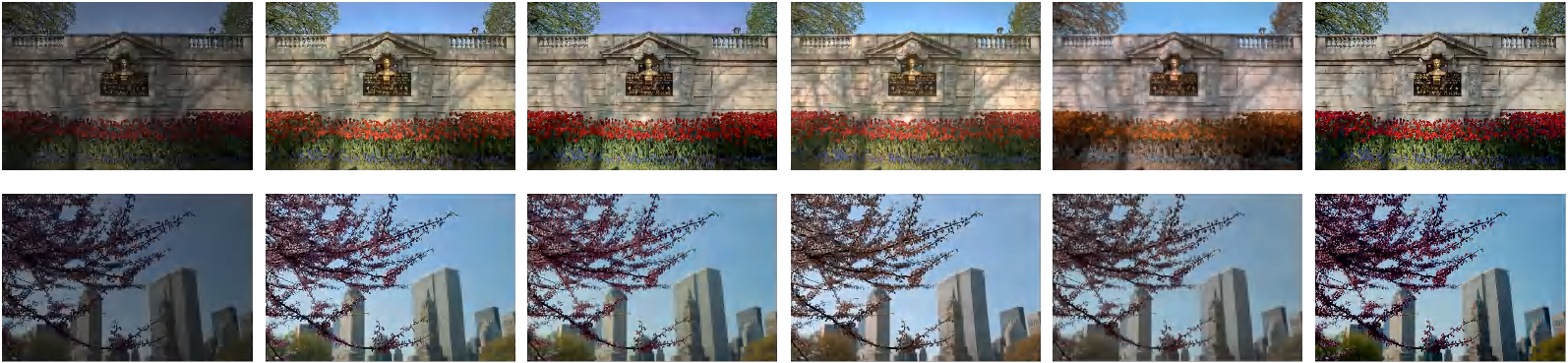}
        \put(7,11){\scriptsize Input}
      \put(21.5,11){\scriptsize $\rm{\overline{H}}_{3}$+$\rm {\overline{H}}_{2}$+$\rm {\overline{H}}_{1}$ }
      \put(38.5,11){\scriptsize $\rm{\overline{H}}_{3}$+$\rm {\overline{H}}_{2}$+$\rm {H}_{1}$}
      \put(55.5,11){\scriptsize $\rm{\overline{H}}_{3}$+$\rm{H}_{2}$+$\rm {H}_{1}$}
      \put(72,11){\scriptsize $\rm{H}_{3}$+$\rm{H}_{2}$+$\rm {H}_{1}$}
      \put(88,11){\scriptsize Ground Truth}
      
     \put(7,-1.5){\scriptsize Input}
      \put(21.5,-1.5){\scriptsize $\rm{\overline{H}}_{3}$+$\rm {\overline{H}}_{2}$+$\rm {\overline{H}}_{1}$}
      \put(38.5,-1.5){\scriptsize $\rm{\overline{H}}_{3}$+$\rm {\overline{H}}_{2}$+$\rm {H}_{1}$}
      \put(55.5,-1.5){\scriptsize $\rm{\overline{H}}_{3}$+$\rm{H}_{2}$+$\rm {H}_{1}$}
      \put(72,-1.5){\scriptsize $\rm{H}_{3}$+$\rm{H}_{2}$+$\rm {H}_{1}$}
      \put(88,-1.5){\scriptsize Ground Truth}
    \end{overpic}
    \vspace{-2mm}
    \caption{\textbf{Visual quality comparison of exposure corrected images processed by our MSLT(1st row) and MSLT+(2nd row) with some high-frequency layers in Laplacian pyramid unprocessed by MSLT/MSLT+}. ``$\rm{{H}}_{i}$": the unprocessed high-frequency layer. ``$\rm{\overline{H}}_{i}$": the exposure-corrected high-frequency layer.}
    \label{fig:high}
    \vspace{-3mm}   
\end{figure*}


\begin{figure*}[h]
    \setlength{\abovecaptionskip}{3pt}
    \centering
    \begin{overpic}[width=1\textwidth]{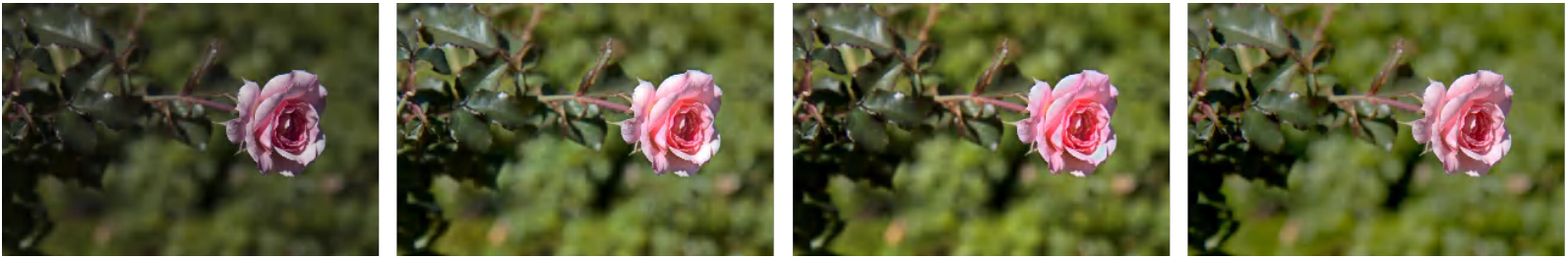}
    \put(9,-1.25){\footnotesize Input}
    \put(34,-1.25){\footnotesize not shared}
    \put(58.5,-1.25){\footnotesize shared (MSLT)}
    \put(83.5,-1.25){\footnotesize Ground Truth}
    \end{overpic}
    \vspace{-2mm}
    \caption{\textbf{Visual quality comparison of exposure corrected images processed by our MSLT with the parameters of $1\times1$ convolutions shared or not.} ``not shared'': deploy independent convolutions between each high-frequency layer. ``shared'': small MLPs in different high-frequency layers share convolution parameters.}
    \label{fig:share}
    \vspace{-3mm}   
\end{figure*}
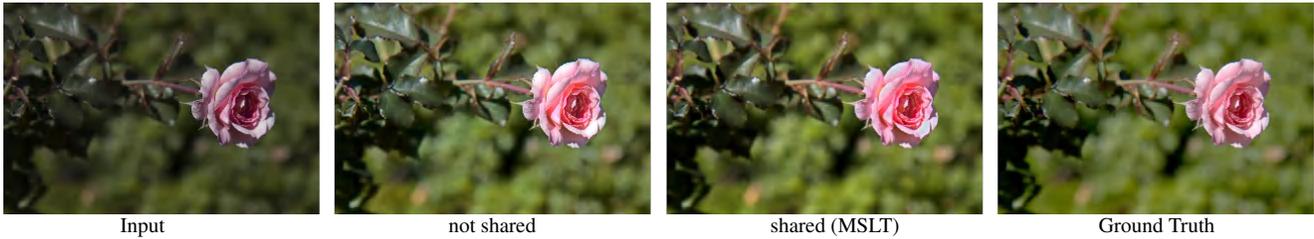


\begin{figure*}[h]
    \setlength{\abovecaptionskip}{3pt}
    \centering
    \begin{overpic}[width=1\textwidth]{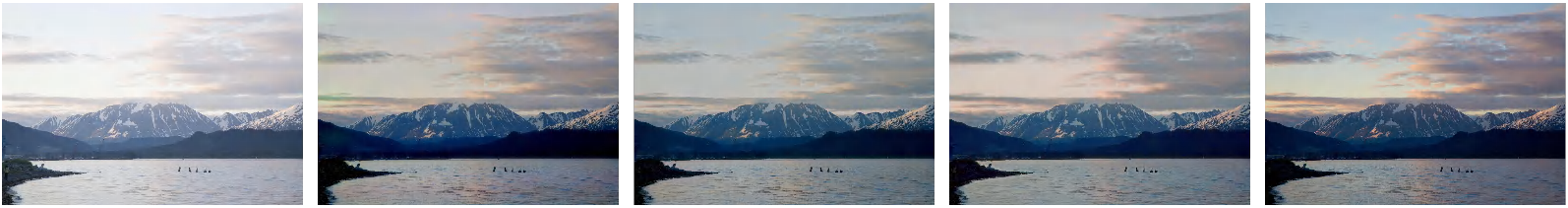}
      \put(8.5,-1.5){\footnotesize Input}
      \put(28,-1.5){\footnotesize GAP}
      \put(48,-1.5){\footnotesize GSP}
      \put(64.5,-1.5){\footnotesize GAP+GSP (MSLT)}
      \put(86,-1.5){\footnotesize Ground Truth}   
    \end{overpic}
    \vspace{-2mm}
    \caption{\textbf{Visual quality comparison of exposure corrected images processed by our MSLT which handles whether or not GAP and GSP are used in CFD moudle.} ``GAP'' (``GSP''): use only ``GAP'' (``GSP'') in our CFD module. ``GAP + GSP'': use the method of adding the``GAP'' and ``GSP'' in our CFD module.}
    \label{fig:GAP}
    \vspace{-3mm}   
\end{figure*}
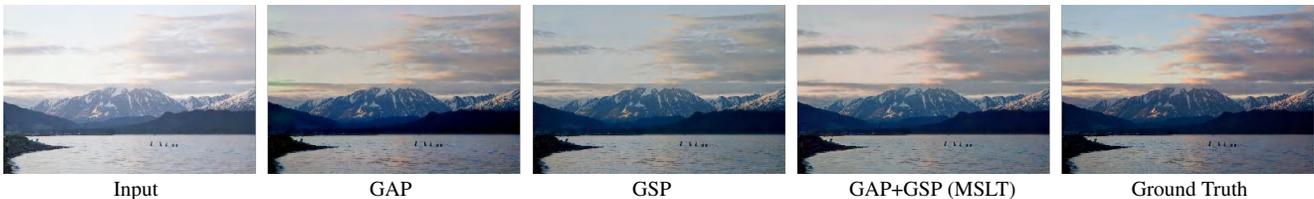


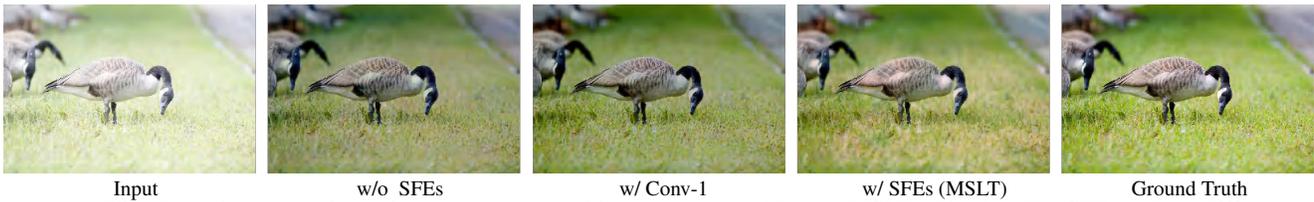
\begin{figure*}[h]
    \setlength{\abovecaptionskip}{3pt}
    \centering
    \begin{overpic}[width=1\textwidth]{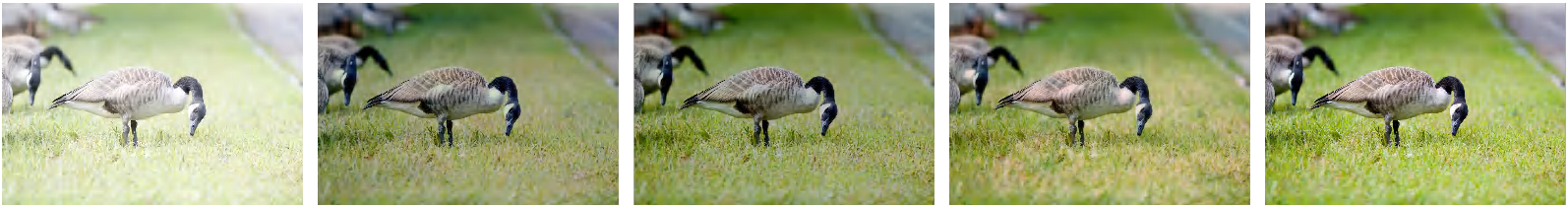}
        \put(8.5,-1.5){\footnotesize Input}
      \put(27,-1.5){\footnotesize w/o \ SFEs}
      \put(47,-1.5){\footnotesize w/ Conv-1}
      \put(65.5,-1.5){\footnotesize w/ SFEs (MSLT)}
      \put(86,-1.5){\footnotesize Ground Truth}
    \end{overpic}
    \vspace{-2mm}
    \caption{\textbf{Visual quality comparison of exposure corrected images processed by our MSLT which handles SFE modules differently.} ``w/o \ SFEs":  SFE moudles are removed from HFD. ``w/ Conv-1": only one convolution and ReLU are left in HFD. `` w/ SFEs": our MSLT. }
    \label{fig:sfe}
    \vspace{-3mm}   
\end{figure*}


\begin{figure*}[h]
    \setlength{\abovecaptionskip}{3pt}
    \centering
    \begin{overpic}[width=1\textwidth]{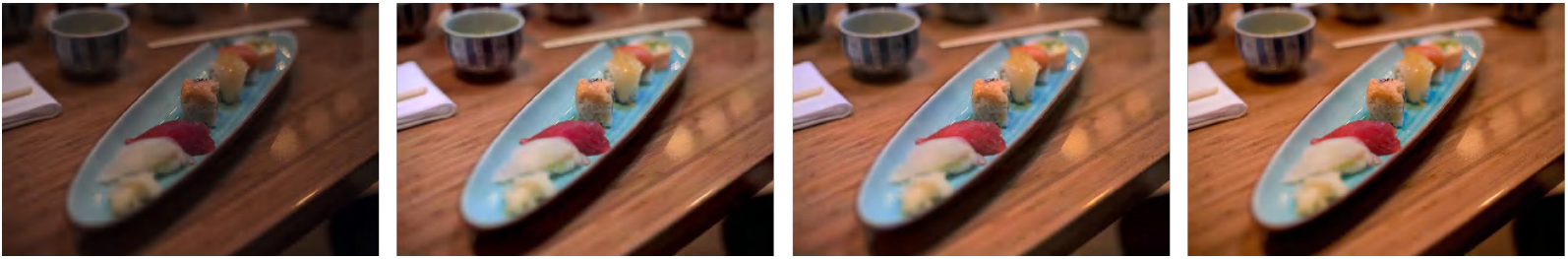}
        \put(8.5,-1.5){\footnotesize Input}
      \put(31,-1.5){\footnotesize Context-aware feature}
      \put(55,-1.5){\footnotesize Residual feature (MSLT)}
      \put(83,-1.5){\footnotesize Ground Truth}
    \end{overpic}
    \vspace{-2mm}
    \caption{\textbf{Visual quality comparison of exposure corrected images processed by our MSLT with different inputs to SFE}. ``Context-aware feature":  the context-aware feature is fed into SFE. ``Residual feature": residual feature is fed into SFE.}
    \label{fig:res}
    \vspace{-3mm}   
\end{figure*}
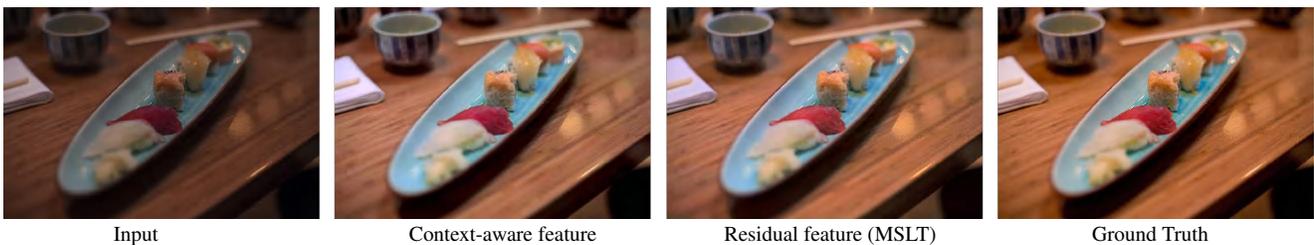

\end{document}